%% file: _main.tex
\documentclass[10pt,twocolumn,letterpaper]{article}

\usepackage{helvet}
\usepackage[pagenumbers]{cvpr}              %

\definecolor{cvprblue}{rgb}{0.21,0.49,0.74}
\usepackage[pagebackref,breaklinks,colorlinks,allcolors=cvprblue]{hyperref}
\usepackage[utf8]{inputenc} %
\usepackage[T1]{fontenc}    %
\usepackage{url}            %
\usepackage{booktabs}       %
\usepackage{amsfonts}       %
\usepackage{nicefrac}       %
\usepackage{microtype}      %
\usepackage[accsupp]{axessibility}  %

\usepackage{amsmath,amssymb,amsthm}
\usepackage{bm}
\usepackage{caption}
\usepackage{subcaption}
\usepackage{graphbox,graphicx}
\usepackage[capitalize,noabbrev,nameinlink,sort]{cleveref}
\usepackage{overpic}

\input{preamble}

\def\paperID{10399} %
\def\confName{CVPR}
\def\confYear{2025}

\title{From Alexnet to Transformers:\\ Measuring the Non-linearity of Deep Neural Networks \\ with Affine Optimal Transport}

\author{%
  Quentin Bouniot$^{\star, 1,2,3,4}$ \quad Ievgen Redko$^{\star,5}$ \quad Anton Mallasto$^{6}$ \quad Charlotte Laclau$^{1}$ \\
  Oliver Struckmeier$^{7}$ \quad Karol Arndt$^{7,8}$ \quad Markus Heinonen$^{9}$ \quad Ville Kyrki$^{7}$ \quad Samuel Kaski$^{9,10}$ \\[5pt]
  $^{1}$LTCI, T\'el\'ecom Paris, Institut Polytechnique de Paris, France, \hspace{0.5pt} $^{2}$Technical University of Munich, \\
  $^{3}$Helmholtz Munich, \quad
  $^{4}$Munich Center for Machine Learning (MCML), \quad 
  $^{5}$Noah's Ark Lab, Paris, \\ $^{6}$Readpeak, \quad  
  $^{7}$Aalto University, Finland, Dept. of Electrical Engineering and Automation, \\ 
  $^{8}$Nomagic, \quad $^{9}$Aalto University, Finland, Department of Computer Science, \\
  $^{10}$University of Manchester, UK, Department of Computer Science
}

\begin{document}
\maketitle
\def\thefootnote{$\star$}\footnotetext{Equal contribution.}
\def\thefootnote{\arabic{footnote}}

\input{sec/0_abstract}    
\input{sec/1_intro}
\input{sec/2_background}
\input{sec/3_contribs}

\input{sec/4_experiments}

\input{sec/5_discussion}

{
    \small
    \bibliographystyle{ieeenat_fullname}
    \bibliography{references}
}

\input{sec/X_suppl}

\end{document}

%% file: preamble.tex
\input{math_commands}

\def\R{{\mathbb{R}}}

\newtheorem{theorem}{Theorem}[section]

\newtheorem{proposition}[theorem]{Proposition}
\newtheorem{corollary}[theorem]{Corollary}

\newtheorem{remark}[theorem]{Remark}

\newcommand{\N}{\mathcal{N}}

\newcommand{\Probs}{\mathcal{P}}
\newcommand{\law}{\mathcal{L}}

\newcommand{\ADM}{\mathrm{ADM}}
\newcommand{\aff}{\mathrm{aff}}

\newcommand{\OT}{\mathrm{OT}}

\newcommand{\Expect}{\operatorname{\mathop{{}\mathbb{E}}}}

%% file: math_commands.tex
\usepackage{amsmath,amsfonts,bm}

\def\eqref#1{equation~\ref{#1}}

\def\1{\bm{1}}

\DeclareMathAlphabet{\mathsfit}{\encodingdefault}{\sfdefault}{m}{sl}
\SetMathAlphabet{\mathsfit}{bold}{\encodingdefault}{\sfdefault}{bx}{n}

\newcommand{\E}{\mathbb{E}}

\newcommand{\R}{\mathbb{R}}

\DeclareMathOperator*{\argmin}{arg\,min}

\DeclareMathOperator{\Tr}{Tr}

%% file: sec/0_abstract.tex
\begin{abstract}
In the last decade, we have witnessed the introduction of several novel deep neural network (DNN) architectures exhibiting ever-increasing performance across diverse tasks. Explaining the upward trend of their performance, however, remains difficult as different DNN architectures of comparable depth and width -- common factors associated with their expressive power -- may exhibit a drastically different performance even when trained on the same dataset. In this paper, we introduce the concept of the non-linearity signature of DNN, the first theoretically sound solution for approximately measuring the non-linearity of deep neural networks. Built upon a score derived from closed-form optimal transport mappings, this signature provides a better understanding of the inner workings of a wide range of DNN architectures and learning paradigms, with a particular emphasis on the computer vision task. We provide extensive experimental results that highlight the practical usefulness of the proposed non-linearity signature and its potential for long-reaching implications. The code for our work is available at \url{https://github.com/qbouniot/AffScoreDeep}.
\end{abstract}

%% file: sec/1_intro.tex
\section{Introduction}
Deep neural networks (DNNs) are undoubtedly the most powerful AI models currently available \citep{lecun2015deep,schmidhuber2015deep,jordan2015machine,goodfellow2016deep,litjens2017survey}. Their performance on many tasks, including natural language processing (NLP) \citep{he2021deberta} and computer vision \citep{he_surpassing}, is already on par or exceeds that of a human being. One of the reasons explaining such progress is of course the increasing computational resources \citep{openai_ai_and_compute_2018,strubell2019energy}. Another one is the endeavour for finding ever more efficient neural architectures pursued by researchers over the last decade. As of today, the transformer architecture \citep{vaswani2017attention} has firmly imposed itself as a number one choice for most, if not all, of the recent breakthroughs \citep{brown2020language,touvron2023llama,OpenAI_GPT4_2023} in machine learning. \\
\noindent \textbf{Limitations \phantom{.}} But why transformers are more capable than other architectures? Answering this question requires finding a meaningful measure to compare the different famous models over time gauging the trend of their intrinsic capacity. For such a comparison to be informative, it is particularly appropriate to consider the computer vision field that produced many of the landmark neural architectures improving upon each other over the years. Indeed, the decade-long revival of deep learning started with Alexnet's \citep{alexnet} architecture, the winner of the ImageNet Large Scale Visual Recognition Challenge \citep{ILSVRC15} in 2012. By achieving a significant improvement over the traditional approaches, Alexnet was the first truly deep neural network to be trained on a dataset of such scale, suggesting that deeper models were likely to bring even more gains. In the following years, researchers proposed novel ways to train deeper models with hundreds of layers \citep{Simonyan15,szegedy2016rethinking,he2016deep,huang2017densely}  pushing the performance frontier even further. The AI research landscape then reached a turning point with the proposal of transformers \citep{vaswani2017attention}, starting their unprecedented dominance first in NLP and then in computer vision \citep{vit_first}. Surprisingly, transformers are not particularly deep, and the size of their landmark vision architecture is comparable to that of Alexnet, and this despite a significant performance gap between the two. Ultimately, this gap should be explained by the differences in the expressive power \citep{survey_uat} of the two models: a term used to denote the ability of a DNN to approximate functions of a certain complexity. Unfortunately, the existing theoretical results related to this either associate higher expressive power with depth \citep{eldan16,safran17a,JMLR:v20:17-612} or width \citep{raghu17a,MontufarPCB14,width_expressive,VardiYS22} falling short in comparing different families of architectures. This, in turn, limits our ability to understand what underpins the achieved progress and what challenges and limitations still exist in the field, guiding future research efforts.  \\
\noindent \textbf{Contributions \phantom{.}} We argue that quantifying the non-linearity of a DNN may be what we were missing so far to understand the evolution of the deep learning models at a more fine-grained level. To verify this hypothesis in practice, we propose a first theoretically sound measure, called the \textit{affinity score}, that estimates the non-linearity of a given (activation) function using optimal transport (OT) theory. We use the proposed affinity score to introduce the concept of the \textit{non-linearity signature} of DNNs defined as a sequence of affinity scores of all its activation functions. \\
\noindent \textbf{Summary of findings \phantom{.}}
\begin{enumerate}
    \item The non-linearity of activation functions depends on the range of input values and its overall shape (Section \ref{sec:aff_score}, \cref{fig:activations}). This impacts the overall complexity of the model even for randomly initialized MLPs (\cref{fig:redshift}).
    \item Until the arrival of vision transformers in 2020, the trend in state-of-the-art vision DNNs was to decrease the average non-linearity of activation functions, rather than to increase it. Vision transformers changed this pattern drastically and are the most non-linear architectures among all landmark ones (Section \ref{sec:experiments}, \cref{fig:timeline} (A)).
    \item Non-linearity signatures of DNNs capture and allow to identify in a meaningful way the similarities in the architectural choices of the different models (Section \ref{sec:experiments}, \cref{fig:timeline} (B), (C)). Non-linearity signature also reflects the exact role of residual connections in increasing the linearity of activation functions, and thus, of the network (Section \ref{sec:experiments}, \cref{fig:resnets}). 
    \item Simple statistics extracted from the non-linearity signatures of DNNs correlate with their performance on Imagenet dataset (Section \ref{sec:experiments}, \cref{fig:correlations_accuracy}). 
    \item Non-linearity signature is unique, as there is no other metric correlating strongly with it across all architectures (Section \ref{sec:experiments}, \cref{tab:correlations}). 
\end{enumerate}

%% file: sec/2_background.tex
\section{Background}\label{sec:background}
\noindent \textbf{Optimal Transport \phantom{.}} 
Let $(X,d)$ be a metric space equipped with a lower semi-continuous \emph{cost function} $c:X\times X \to \mathbb{R}_{\geq 0}$, e.g the Euclidean distance $c(x,y) = \|x-y\|$. Then, the Kantorovich formulation of the OT problem between two probability measures $\mu, \nu \in \Probs(X)$ is given by 
\begin{equation}\label{eq:kantorovich}
    \OT_c(\mu, \nu) = \min_{\gamma\in \ADM(\mu,\nu)}\E_\gamma[c],
\end{equation}
where $\ADM(\mu,\nu)$ is the set of joint probabilities with marginals $\mu$ and $\nu$, and $\E_\nu[f]$ denotes the expected value of $f$ under $\nu$. The optimal $\gamma$ minimizing \eqref{eq:kantorovich} is called the \emph{OT plan}. Denote by $\law(X)$ the law of a random variable $X$. Then, the OT problem extends to random variables $X,Y$ and we write $\OT_c(X,Y)$ meaning $\OT_c(\law(X), \law(Y))$.

Assuming that either of the considered measures is \emph{absolutely continuous}, then the Kantorovich problem is equivalent to the \emph{Monge problem}
\begin{equation}
    \OT_c(\mu,\nu) = \min\limits_{T: T_\#\mu = \nu} \E_{X\sim \mu}[c(X,T(X))],
\end{equation}
where the unique minimizing $T$ is called the \emph{OT map}, and $T_\#\mu$ denotes the \emph{push-forward measure}, which is equivalent to the \emph{law} of $T(X)$, where $X\sim \mu$. \\  
\noindent \textbf{Wasserstein distance \phantom{.}} Let $X$ be a random variable over $\R^d$ satisfying $\E[\|X-x_0\|^2]<\infty$ for some $x_0\in \R^d$, and thus for any $x\in \R^d$. We denote this class of random variables by $\Probs_2(\R^d)$. Then, the $2$-Wasserstein distance $W_2$ between $X,Y\in \Probs_2(\R^d)$ is defined as
\begin{equation}
    W_2(X,Y) = \OT_{||x-y||^2}(X, Y)^{\frac{1}{2}}.
\end{equation}
We now proceed to the presentation of our main contribution. 

%% file: sec/3_contribs.tex
\begin{figure*}
    \centering
    \includegraphics[width=.85\linewidth]{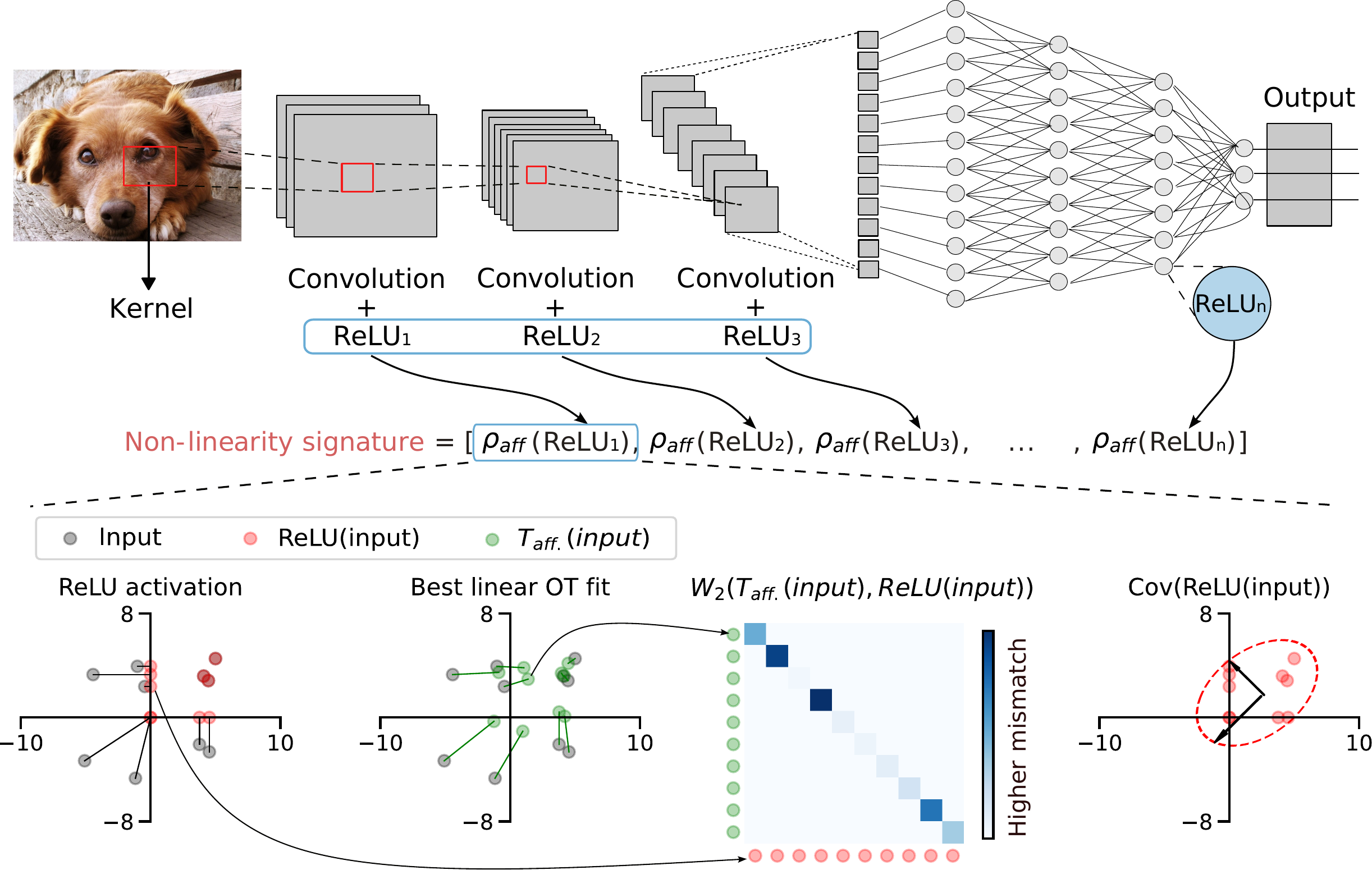}
    \caption{Illustration of how the non-linearity of a given neural network is measured. (\textbf{Top}) The non-linearity signature of a DNN is a collection of affinity scores calculated for each activation function spread across its hidden layers. (\textbf{Bottom}) The affinity score is calculated based on 3 main steps. First, given an input (grey) and an output (red) of an activation function (\emph{left}), we estimate the best affine OT fit $T_\mathrm{aff}(X)$ (green) transporting the input to the output (\emph{middle-left}). Second, we measure the mismatch between the two by summing the transportation costs (\emph{middle-right}) to obtain the Wasserstein distance $W_2(T_\aff X,Y)$. Finally, this distance is normalized with the magnitudes of variance (arrows in the rightmost plot) of the output data based on its covariance matrix.}
    \label{fig:non-linearity_signature}
\end{figure*}

\section{Non-linearity signature of DNNs}\label{sec:aff_score}
Among all non-linear operations introduced into DNNs in the last several decades, activation functions remain the only structural piece that they all inevitably share. For instance, without non-linear activation functions, popular multi-layer perceptrons, no matter how deep, reduce to a linear function unable to learn complex patterns.
Activation functions were also early identified \citep{hornik1989multilayer,barron,Kurt1991251,cybenko} as a key to making even a shallow network capable of approximating any function, however complex it may be, to arbitrary precision. 

We thus build our study on the following intuition: if activation functions play an important role in making DNNs non-linear, then measuring their degree of non-linearity can provide us with an approximation of the DNN's non-linearity itself. To implement this intuition in practice, however, we first need to find a way to measure the non-linearity of an activation function. Surprisingly, there is no widely accepted measure for this, neither in the field of mathematics nor in the field of computer science. To fill this gap, we will use the OT theory to develop a so-called \textit{affinity score} below.

\subsection{Affinity score}

\textbf{Identifiability \phantom{.}} We consider the pre-activation signal $X$ of an activation function within a neural network, and the post-activation signal $\sigma(X)$ denoted by $Y$ as input and output random variables. Our first step to build the affinity score then is to ensure that we can identify when $\sigma$ is linear with respect to (wrt) $X$ (for instance, when an otherwise non-linear activation is \emph{locally linear} at the support of $X$). To show that such an identifiability condition can be satisfied with OT, we first recall the following classic result from the literature characterizing the  OT maps.
\begin{theorem}[\citep{smith1987note}]\label{thm:OT_maps}
Let $X\in \Probs_2(\R^d)$, $T(x) = \nabla \phi(x)$ for a convex function $\phi$ with $T(X)\in\Probs_2(\R^d)$. Then, $T$ is the unique optimal OT map between $\mu$ and $T_\#\mu$.
\end{theorem}
Using this theorem about the uniqueness of OT maps expressed as gradients of convex functions, we can prove the following result (all proofs are in Appendix \ref{subsec:proofs}):  
\begin{corollary}\label{cor:affine_ot_map}
Without loss of generality, let $X,Y\in \Probs_2(\R^d)$ be centered, and let $Y= \sigma(X) = TX$, where $T$ is a positive definite linear transformation. Then, $T$ is the OT map from $X$ to $Y$.
\end{corollary}
Whenever the activation function $\sigma$ is linear, the solution to the OT problem $T$ exactly reproduces it. 
\medskip

\noindent \textbf{Characterization \phantom{$\frac{1}{2}$}} We now seek to understand whether $T$ can be characterized more explicitly. For this, we prove the following theorem stating that $T$ can be computed in closed-form using the normal approximations of $X$ and $Y$.
\begin{theorem}\label{thm:affine_OT}
Let $X,Y\in \Probs_2(\R^d)$ be centered and $Y = TX$ for a positive definite matrix $T$. Let  $N_X \sim \N(\mu(X),\Sigma(X))$ and $N_Y \sim \N(\mu(Y),\Sigma(Y))$ be their normal approximations where $\mu$ and $\Sigma$ denote mean and covariance, respectively. Then, $W_2(N_X,N_Y) = W_2(X,Y)$
and $T = T_\aff$, where $T_\aff$ is the OT map between $N_X$ and $N_Y$ and can be calculated in closed-form
    \begin{equation}\label{eq:affine_map}
        \begin{aligned}
            &T_\aff(x) = Ax + b,\\
            &A = \Sigma(Y)^\frac{1}{2}\left(\Sigma(Y)^\frac{1}{2}\Sigma(X)\Sigma(Y)^\frac{1}{2}\right)^{-\frac{1}{2}}\Sigma(Y)^\frac{1}{2}\\
            &b = \mu(Y) - A\mu(X).
        \end{aligned} 
    \end{equation}
\end{theorem}
Theorem 3.3 is reminiscent of the classical statement about the 2-Wasserstein distance $W_2$ between two normal distributions \citep{cot_peyre_cutu} but is significantly more general. The statement is for arbitrary random variables $X$ and $Y$ that satisfy $Y=T(X)$ for a symmetric positive-definite linear transformation $T$. The classical result about normal distributions is then a direct corollary.
\medskip

\noindent \textbf{Upper bound \phantom{.}} When the activation $\sigma$ is non-linear wrt $X$, the affine OT mapping $T_\mathrm{aff}(X)$ will deviate from the true activation outputs $Y$. One important step toward quantifying this deviation is given by the famous Gelbrich bound, formalized by means of the following theorem:
\begin{theorem}[Gelbrich bound~\citep{gelbrich1990formula}]\label{thm:gaussians_lower_bound_2was}
Let $X,Y\in \Probs_2(\R^d)$ and let $N_X,N_Y$ be their normal approximations. Then, $W_2(N_X,N_Y) \leq W_2(X,Y)$.
\end{theorem}
This upper bound provides a first intuition of why OT can be a great tool for measuring non-linearity: the cost of the affine map solving the OT problem on the left-hand side increases when the map becomes non-linear. We now upper bound the difference between $W_2(N_X, N_Y)$ and $W_2(X,Y)$, two quantities that coincide \textit{only} when $\sigma$ is linear.
\begin{proposition}\label{prop:affine_vs_standard_OT}
Let $X,Y\in \Probs_2(\R^d)$ and $N_X, N_Y$ be their normal approximations. Then, 
\begin{enumerate}
    \item $
    \left| W_2(N_X, N_Y) - W_2(X,Y)\right| \leq \frac{2\Tr\left[\left(\Sigma(X)\Sigma(Y)\right)^\frac{1}{2}\right]}{\sqrt{\Tr[\Sigma(X)] + \Tr[\Sigma(Y)]}}.$
    \item For $T_\aff$ as in (\ref{eq:affine_map}), $W_2(T_\aff X,Y) \leq \sqrt{2\Tr\left[\Sigma(Y)\right]}.$ 
\end{enumerate}
\end{proposition}
To have a more informative non-linearity measure, we now need to normalize the non-negative Wasserstein distance $W_2(T_\aff X,Y)$ to an interpretable interval of $[0,1]$. The bound given in Proposition~\ref{prop:affine_vs_standard_OT} lets us define the following \emph{affinity score}
\begin{equation}
    \rho_\aff(X,\sigma(X)) = 1 - \frac{W_2(T_\aff X,\sigma(X))}{\sqrt{2\Tr[\Sigma(\sigma(X))]}}.
    \label{eq:aff_score}
\end{equation}
The proposed affinity score quantifies how far a given activation $\sigma$ is from an affine transformation. It is equal to 1 for any input for which the activation function is linear, and 0 when it is maximally non-linear, i.e., when $T_\aff X$ and $\sigma(X)$ are independent random variables. We illustrate the computation of the affinity score in \cref{fig:non-linearity_signature}.
\begin{remark} One may wonder whether a simpler alternative to the affinity score can be to use, instead of $T_\aff$, a mapping $T_W(x) = Wx$ defined as a solution of a linear regression problem $\min_W ||Y - WX||_F^2$. Then, one can use the coefficient of determination ($R^2$ score) to measure how well $T_W$ fits the observed data. This approach, however, has two drawbacks. First, following the famous Gauss-Markov theorem, $T_W$ is an optimal \textit{linear} (linear in $Y$) estimator. On the contrary, $T_\aff $ is a globally optimal non-linear mapping aligning $X$ and $Y$. Second, $R^2$ compares the fit of $T_W$ with that of a mapping outputting $\mu(Y)$ for any value of $X$. This is contrary to $\rho_\aff$ that compares how well $T_\aff$ fits the data wrt to the worst possible cost incurred by $T_\aff $ as quantified in Proposition \ref{prop:affine_vs_standard_OT}. This gives us a bounded score, i.e. $\rho_\aff \in [0,1]$, whereas $R^2$ is not lower bounded, i.e. $R^2 \in [-\infty,1]$. We confirm experimentally in Section \ref{sec:experiments} that the two coefficients do not correlate consistently across the studied DNNs suggesting that $R^2$ is a poor proxy to $\rho_\aff$.
\end{remark}

\subsection{Non-linearity and activation functions}

\paragraph{What makes an activation function non-linear?} 
We now want to understand the mechanism behind achieving a lower or higher non-linearity with a given (activation) function. This will explain what the different values of the affinity scores stand for when studying trained DNNs. In \cref{fig:activations} (A), we show how the ReLU function \citep{nair2010rectified}, defined element-wise as $\text{ReLU}(x) = \text{max}(0,x)$, achieves its varying degree of non-linearity. Interestingly, this degree depends only on the range of the input values. %
Second, in \cref{fig:activations} (B) we also show how the shape of activation functions impacts their non-linearity for a fixed input: surprisingly, piece-wise linear ReLU function is more non-linear than $\text{Sigmoid}(x) = 1/(e^{-x}+1)$ \citep{rumelhart1986learning} or $\text{Tanh}(x) = (e^{-x}-e^{x})/(e^{-x}+e^{x})$. Similar observations also apply to compare polynomials of varying degrees (\cref{fig:activations} (C)). We include more visualizations of the affinity score of popular activation functions in \cref{sec:appendix_activations}.

\begin{figure}[!t]
    \centering
    \begin{picture}(\linewidth,0)%
    \put(0, -10){(A)}
    \put(0,-80){(B)}
    \put(0,-135){(C)}
    \end{picture}
    \includegraphics[width=.7\linewidth]{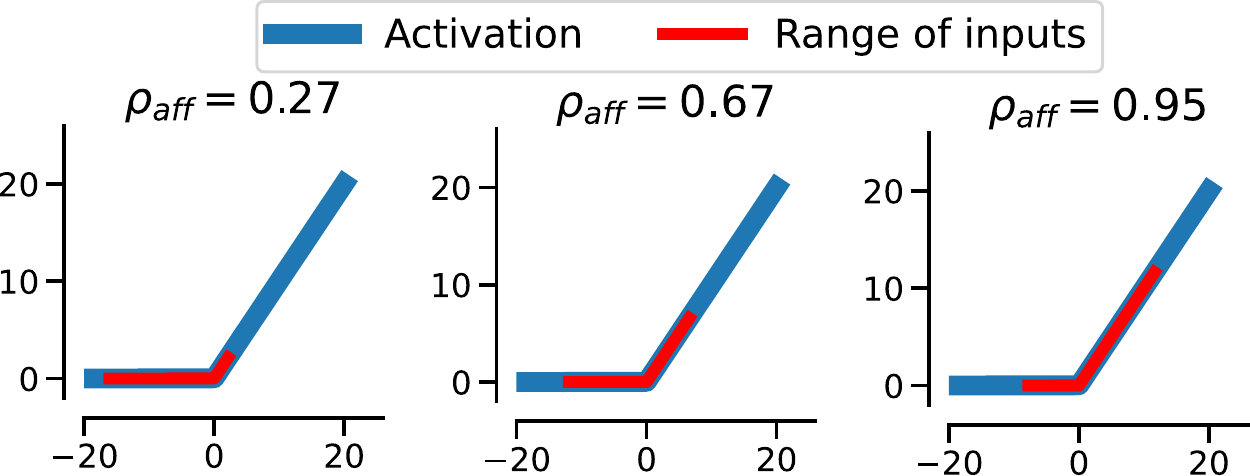}
    \medskip
    
    \includegraphics[width=.82\linewidth]{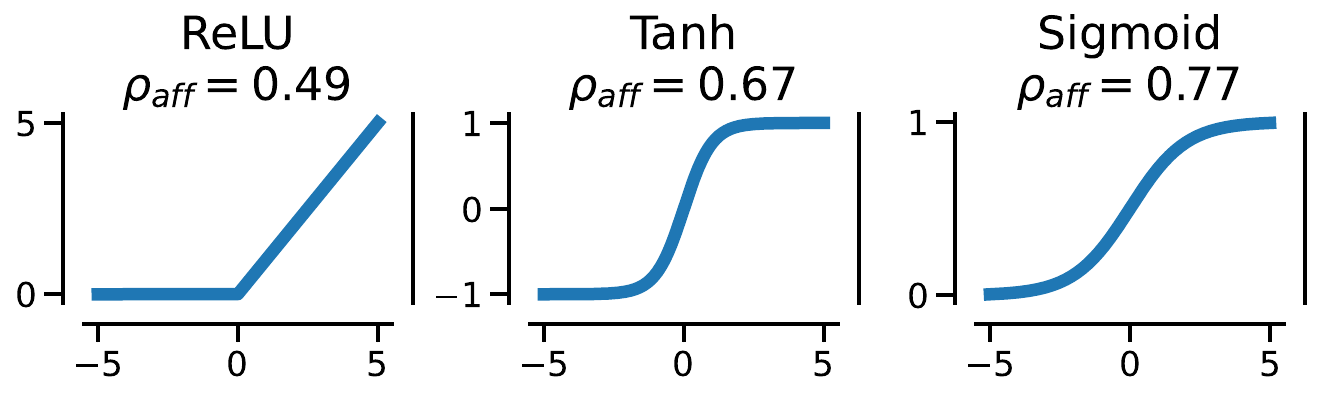}
    \medskip
    
    \includegraphics[width=\linewidth]{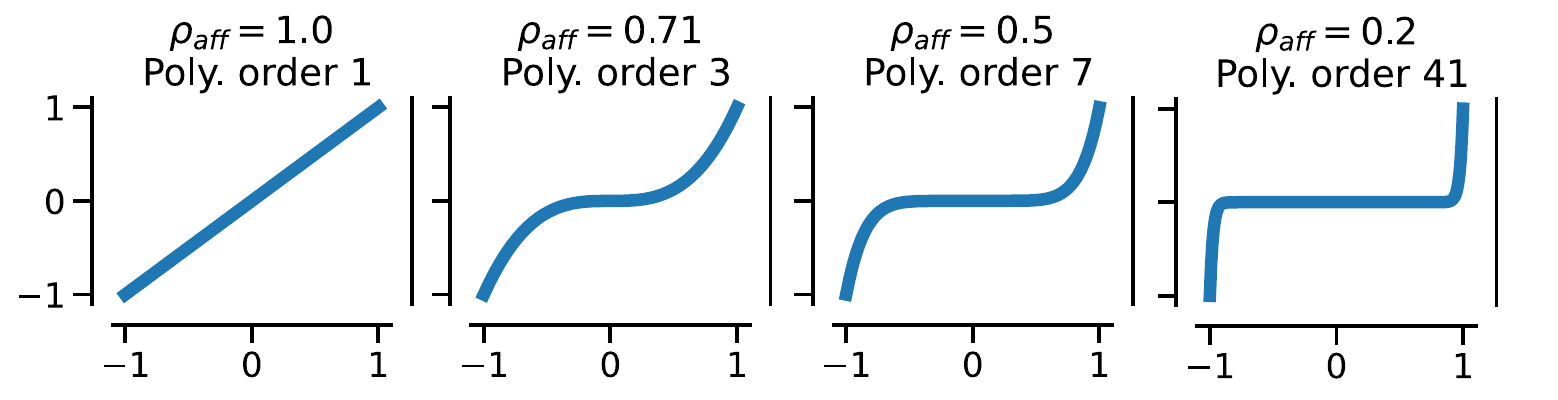}
    \caption{\textbf{(A)} Non-linearity of activation functions depends on the range of input values (\emph{red}), illustrated here on ReLU; \textbf{(B)} ReLU, Tanh, and Sigmoid exhibit different degrees of non-linearity for the same input; \textbf{(C)} Affinity score captures the increasing non-linearity of polynomials of different degrees.}
    \label{fig:activations}
\end{figure}
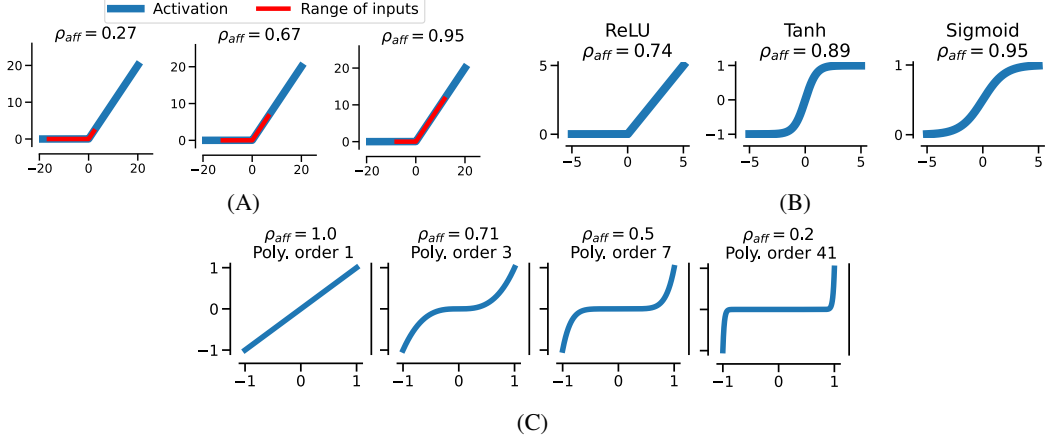
\begin{figure}[!t]
    \centering
    \includegraphics[width=\linewidth]{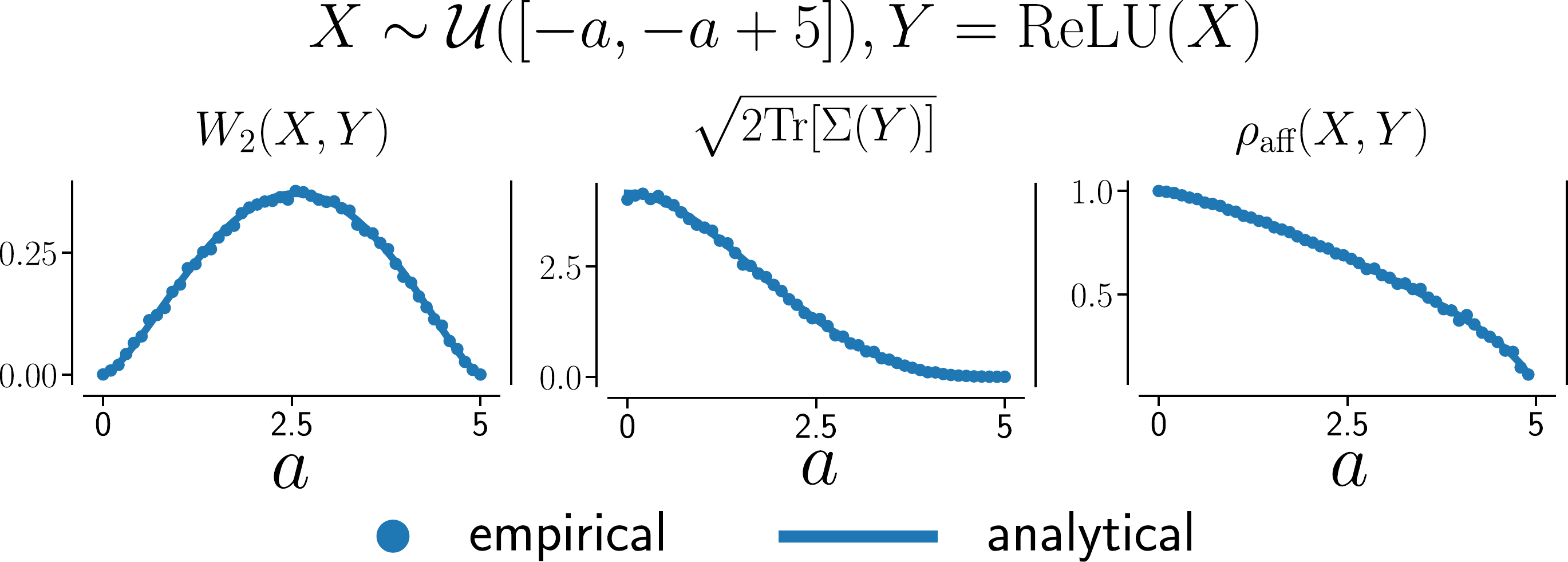}
    \caption{The behavior of \textbf{(Left)} $W_2(X,Y)$ term, \textbf{(Middle)} $\sqrt{2\Tr[\Sigma(Y)]}$ and \textbf{(Right)} $\rho_\text{aff}(X,Y)$.}
    \label{fig:assymetry}
\end{figure}

\medskip

\begin{figure}[!t]
    \centering
    \begin{picture}(\linewidth,0)%
    \put(0, -12){(A)}
    \put(.51\linewidth,-12){(B)}
    \end{picture}
    \includegraphics[width=.49\linewidth]{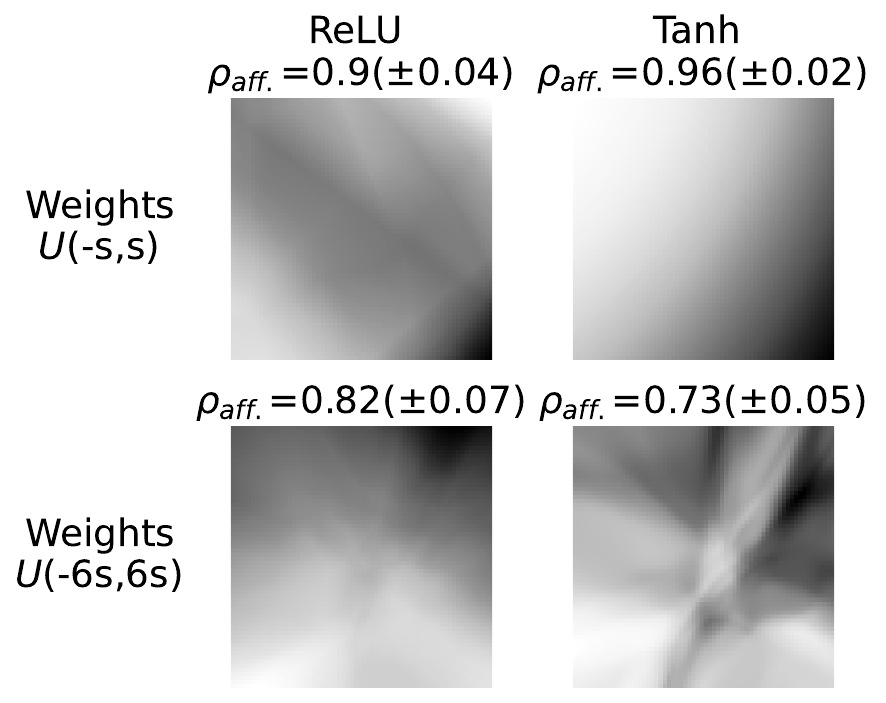}
    \includegraphics[width=.49\linewidth]{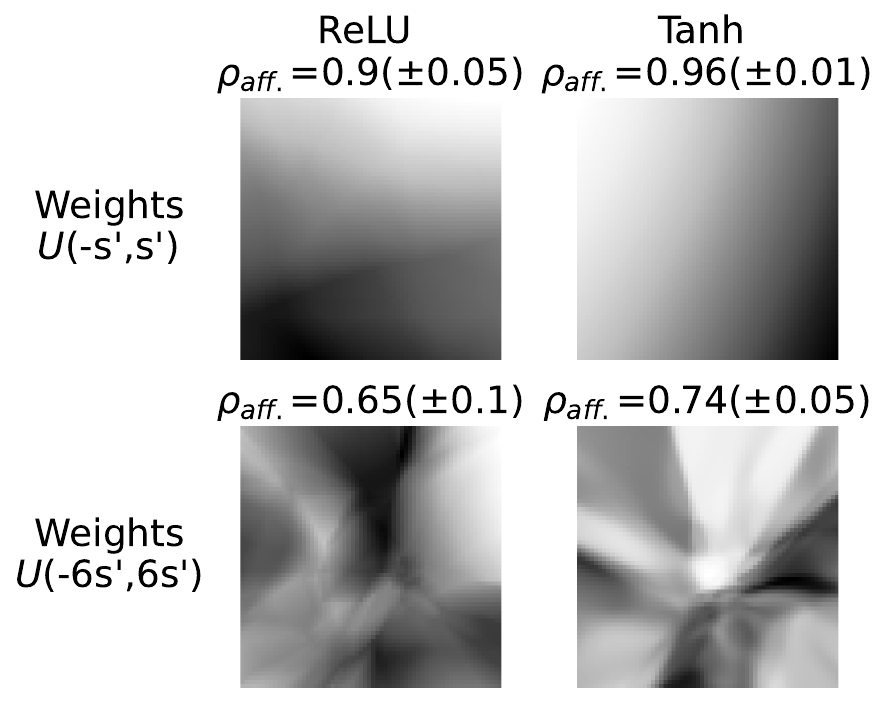}
    \caption{Revisiting the neural redshift phenomenon. \textbf{(A)} MLP with 3 hidden layers \citep{neural_redshift} equipped with either ReLU or Tanh activation functions; \textbf{(B)} Same MLP initialized by setting $s'=s-0.05$ in Glorot initialization. By changing the domain of ReLU, we see that its simplicity bias with changing weight magnitudes vanishes.}
    \label{fig:redshift}
\end{figure}

\noindent \textbf{A perfect non-linearity and asymmetry \phantom{.}} In \cref{fig:activations} (A), we showed that the affinity score is not symmetric around the origin. This is mainly due to the difference in the behavior of the two terms used to calculate $\rho_\text{aff}$. To explain this asymmetry, we derive an analytical expression for $W_2(X,Y)$ and $\sqrt{\Tr[\Sigma(Y)]}$ terms for a special case when $X \sim \mathcal{U}([b, a]), Y = \text{ReLU}(X)$ for $b<0$ and $a>0$ (see \cref{subsec:1d_relu}) and illustrate it for $b=-a$ and $a=-a+5$ for $a \in [0,5]$ in \cref{fig:assymetry}. We note that the numerator -- illustrated in the left part of the plot -- is roughly symmetric and the main source of asymmetry comes from the denominator term (middle part) that captures the amount of variance in the output of the activation function. The affinity score (right part) in 1D then approaches 0 only when both terms tend to 0: a situation where the activation function receives only negative values. Surprisingly, this shows that in our formalism, the ``perfect'' non-linearity doesn't exist. It would be interesting to study whether this holds in higher dimensions.  \\
\noindent \textbf{Neural redshift revisited \phantom{.}}As a follow-up to the previous experiment, we now revisit a recent work by \cite{neural_redshift} that studied the complexity biases carried by the different activation functions in randomly initialized DNNs. The conclusion reached by the authors is that ReLU and its variations have a strong bias toward low complexity which is unaffected by the change in the magnitude of weights when compared, for instance, to tanh. We follow the protocol of \cite{neural_redshift} and consider an MLP with 3 hidden layers and scalar output initialized using Glorot initialization $\mathcal{U}(-s,s)$ \cite{pmlr-v9-glorot10a} for weight matrices and $\mathcal{U}(-1,1)$ for biases. The inputs to the network are $64^2$ evenly spaced 2D coordinates on the grid in $[-1,1]^2$ so that the network output $64^2$ scalars that can be visualized as a grayscale image. In \cref{fig:redshift} (A), we present the results from \cite{neural_redshift} reproducing their claim regarding the comparison between ReLU and tanh when using the different weight magnitudes in the Glorot initialization. This indeed shows that ReLU visually appears to lead to a much simpler function that is independent of the weights' magnitudes. The appearance, however, seems misleading as the average affinity scores of ReLU activations in the considered MLP decrease slightly in the case of higher magnitude weights, yet remain higher than that of tanh. However, our experiments in \cref{fig:activations} suggest that this behaviour should not be universal and that the domain of the activation function should have a strong influence on its complexity. To verify this, we slightly change the $s$ parameter in Glorot initialization by setting it to $s'=s-0.05$. We redo the experiment as before and plot the obtained results in \cref{fig:redshift} (B). It is now apparent that both visually, and quantitatively, the ReLU activations became much more complex within the considered random MLP with the visualization of the function approximating the considered networks becoming almost indistinguishable between ReLU and tanh. Our proposed affinity score captures this change of complexity and provides a more fine-grained quantitative measure for it. \\
\noindent \textbf{Non-linearity signature \phantom{.}}
We now turn our attention to the definition of a non-linearity signature of deep neural networks. We define a neural network $\text{N}$ as a composition of layers $F_i$ where each layer $F_i$ is a function taking as input a tensor $\mathrm{X}_i \in \R^{h_i \times w_i \times c_i}$ (for instance, an image of size $224\times 224 \times 3$ for $i=1$)  and outputting a tensor $\mathrm{Y}_i \in \R^{h_{i+1}\times w_{i+1} \times c_{i+1}}$ used as an input of the following layer $F_{i+1}$. This defines $\text{N} = F_L \odot ... \odot F_i \ ... \odot F_1 = \bigodot_{k=1,\dots, L} F_k$ where $\odot$ stands for a composition.
Non-linearity signature, illustrated in \cref{fig:non-linearity_signature} (top), associates to each network $\text{N}$ a vector of affinity scores calculated over the inputs and outputs of all activation functions across layers.

\subsection{Related work} 
\noindent \textbf{Layer-wise similarity analysis of DNNs \phantom{.}} A line of work that can be distantly related to our main proposal is that of quantifying the similarity of the hidden layers of the DNNs as proposed  \cite{raghuCCA} and \cite{kornblithCKA} (see \citep{DavariHNLWB23} for a complete survey of the subsequent works). \cite{raghuCCA} extracts activation patterns of the hidden layers in the DNNs and use CCA on the singular vectors extracted from them to measure how similar the two layers are. Their analysis brings many interesting insights regarding the learning dynamics of the different convnets, although they do not discuss the non-linearity propagation in the convnets, nor do they propose a way to measure it. \cite{kornblithCKA} proposed to use a normalized Frobenius inner product between kernel matrices calculated on the extracted activations of the hidden layers and argued that such a similarity measure is more meaningful than that proposed by \cite{raghuCCA}. \\
\noindent \textbf{Impact of activation functions \phantom{.}} \cite{survey_activations} provides the most comprehensive survey on the activation functions used in DNNs. Their work briefly discusses the non-linearity of the different activation functions suggesting that piecewise linear activation functions with more linear components are more non-linear (e.g., ReLU vs. ReLU6). The choice of activation function can lead to different implicit biases carried by the DNNs \cite{neural_redshift}. \cite{hayou19a} show theoretically that smooth versions of ReLU allow for more efficient information propagation in DNNs with a positive impact on their performance.  Our work provides a first extensive comparison of all popular activation functions; we also show that smooth version of ReLU exhibit wider regions of high non-linearity (see \cref{sec:appendix_activations}). Furthermore, we illustrate that the choice of activation function by itself doesn't have a determining effect on the inductive bias of a given randomly initialized DNN. A simple rescaling of the weights obtained either as a result of training or through a different initialization strategy, changes the non-linearity of the activation function within a DNN and, subsequently, its inductive bias (see \cref{fig:redshift}). \\
\noindent \textbf{Non-linearity measure \phantom{.}} \cite{philipp2021} proposed the non-linearity coefficient in order to predict the train and test error of DNNs. Their coefficient is defined as a square root of the Jacobian of the neural network calculated wrt its input, multiplied by the covariance matrix of the Jacobian, and normalized by the covariance matrix of the input. The presence of the Jacobian in it calls for the differentiability assumption making its application to most of the neural networks with ReLU non-linearity impossible as is. The authors didn't provide any implementation of their coefficient and we were not able to find any prior study reproducing their results.

%% file: sec/4_experiments.tex
\begin{figure*}[!h]
    \centering
    \begin{picture}(\linewidth,0)%
    \put(0,-10){(A)}
    \put(.61\linewidth,-10){(B)}
    \put(0,-160){(C)}
    \end{picture}
    \includegraphics[width=\linewidth]{figures/pairwise_archs_and_timeline.pdf}
    \includegraphics[width=.9\linewidth]{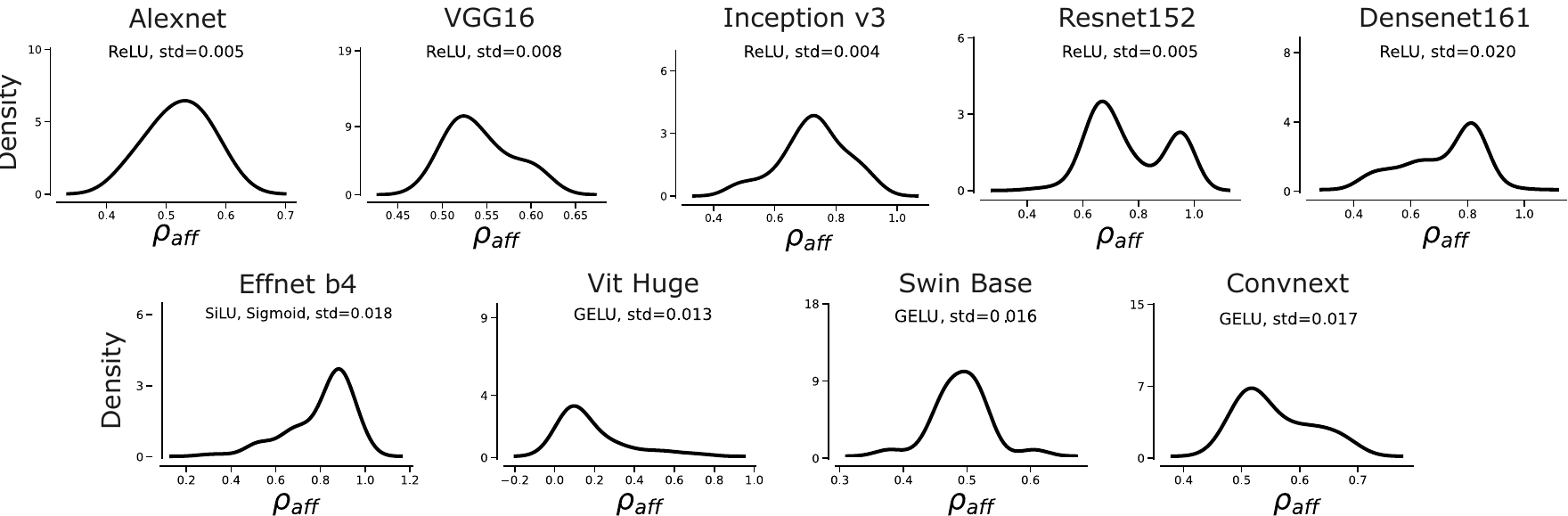}
    \caption{\textbf{(A)} Median, minimum, and maximum values of non-linearity signatures of the different architectures spanning a decade (2012-2022) of computer vision research. \textbf{(B)-(C)} Pairwise similarities and densities of affinity scores' distributions of representative architectures. %
    } 
    \label{fig:timeline}
\end{figure*}

\section{Experimental evaluations}\label{sec:experiments}
We consider computer vision models trained and evaluated on the same Imagenet dataset with 1,000 output categories (Imagenet-1K) publicly available at \citep{torchvision2016}. The non-linearity signatures of different studied models presented in the paper is calculated by passing batches of size 512 through the pre-trained models for the entirety of the Imagenet-1K validation set (see \cref{sec:appendix_datasets} for more datasets) with a total of 50,000 images. We include the following landmark architectures in our study: Alexnet \citep{alexnet}, four VGG models \citep{Simonyan15}, Googlenet \citep{szegedy2014going}, Inception v3 \citep{szegedy2016rethinking}, five Resnet models \citep{he2016deep}, four Densenet models \citep{huang2017densely}, four MNASNet models \citep{Tan_2019_CVPR}, four EfficientNet models \citep{tan19a}, five ViT models, three Swin transformer \citep{liu2021Swin} and four Convnext models \citep{liu2022convnet}. We include MNASNet and EfficientNet models as prominent representatives of the neural architecture search approach \citep{elsken2019neural}. Such models are expected to explicitly maximize the accuracy for a given computational budget. Swin transformer and Convnext models are introduced as ViTs with traditional computer vision priors. Their presence will be useful to better grasp how such priors impact ViTs. We refer the reader to \cref{ax:impl_details} for more practical details. \\
\begin{figure*}[!h]
    \centering
    \includegraphics[width=.33\linewidth]{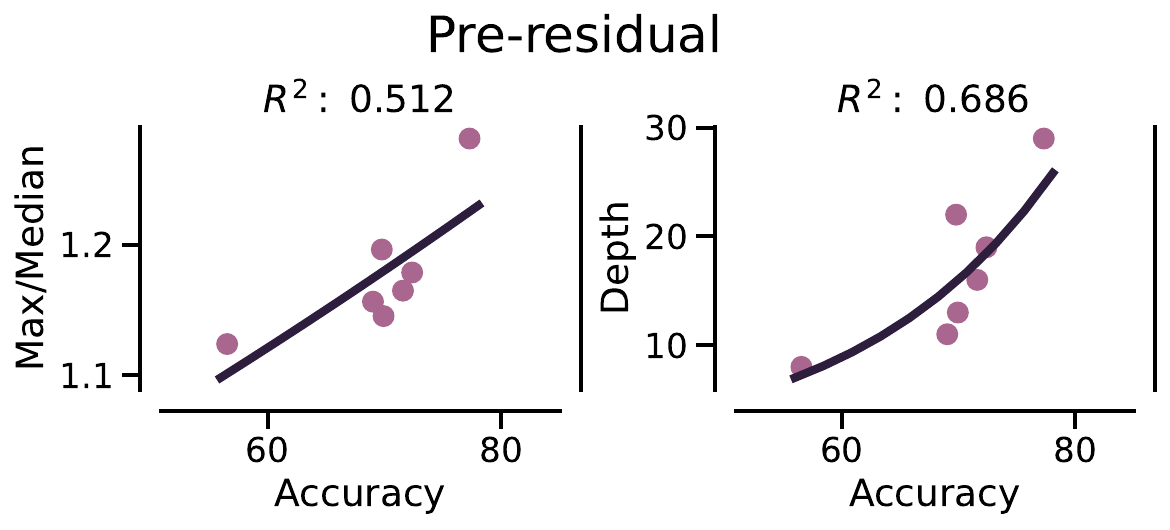}
    \includegraphics[width=.33\linewidth]{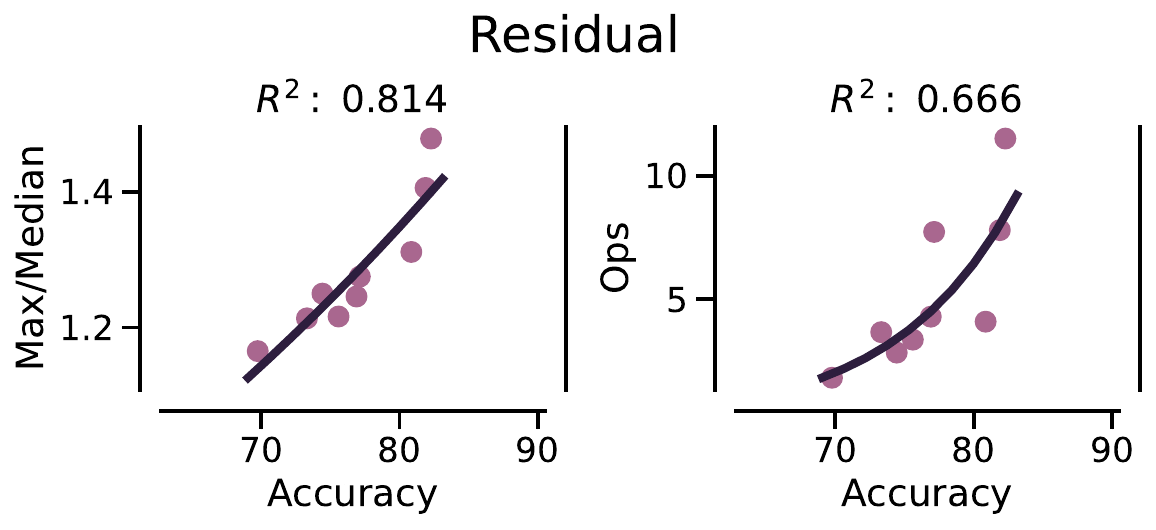}
    \includegraphics[width=.33\linewidth]{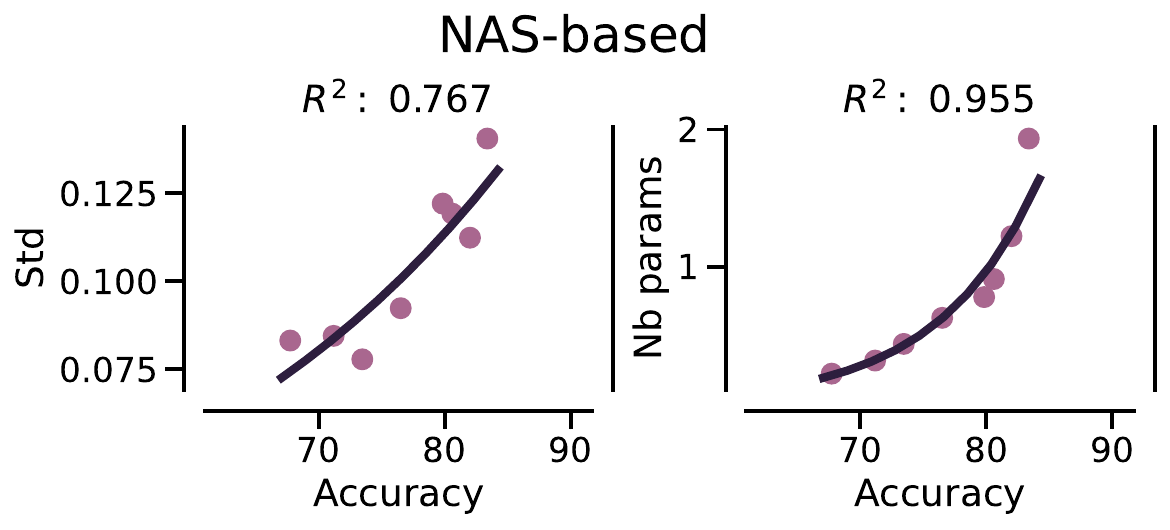}
     \hfill
    \includegraphics[width=.33\linewidth]{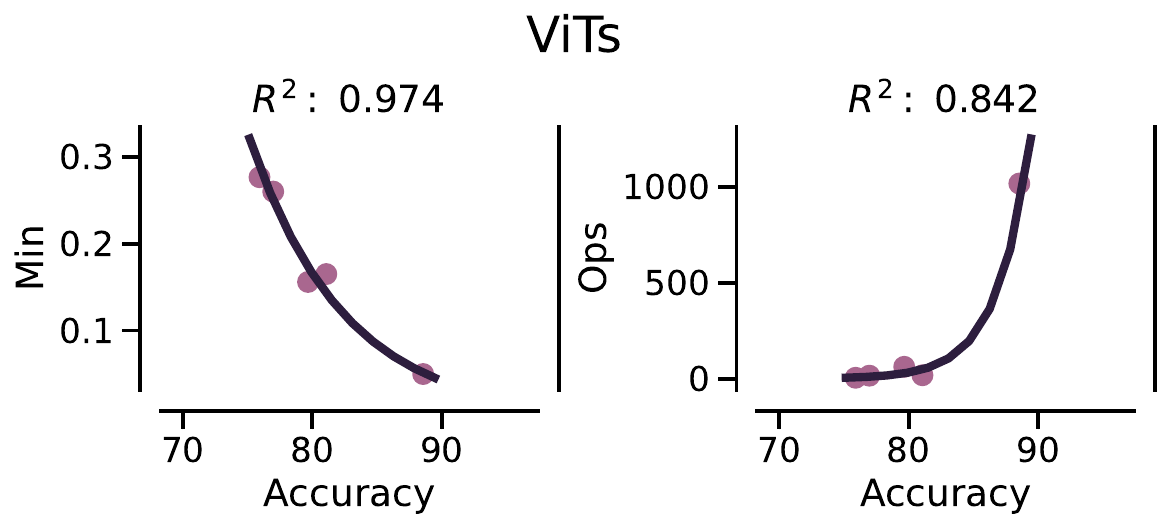}
    \includegraphics[width=.33\linewidth]{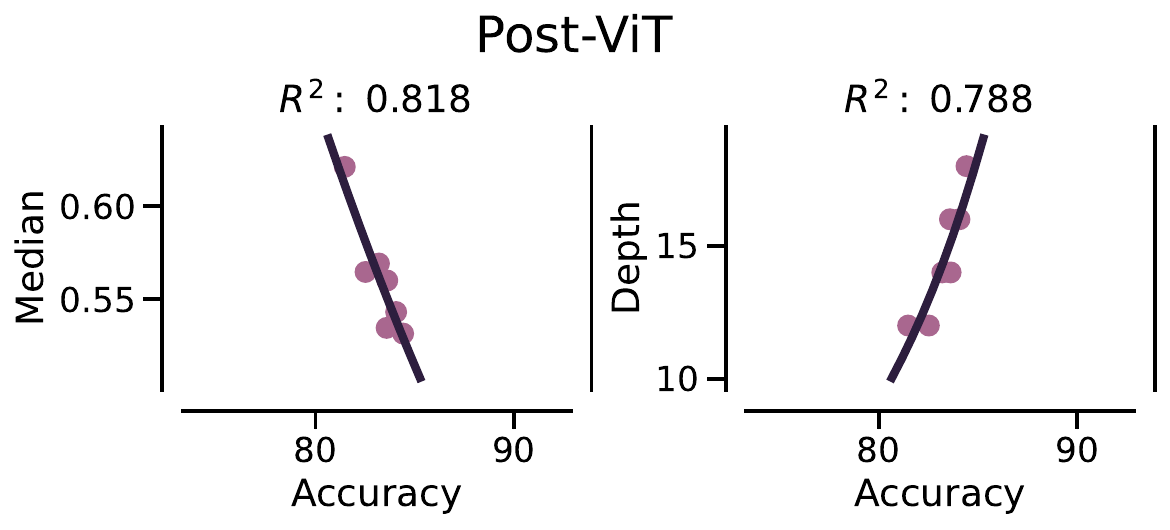}
    \caption{Best found dependency between the different statistics extracted from the non-linearity signatures of the DNN families and their respective Imagenet-1K accuracy. The results are compared in terms of the $R^2$ score against the most precise of the other common DNN characteristics such as depth, size, and the GFLOPS.}
    \label{fig:correlations_accuracy}
\end{figure*}
\noindent \textbf{History of deep vision models at a glance \phantom{.}}
We give a general outlook of the developments in computer vision over the last decade when seen through the lens of their non-linearity. %
In \cref{fig:timeline} (A) we present the minimum, median, and maximum values of the affinity scores calculated for the considered neural networks (see \cref{ax:raw_signatures} for raw non-linearity signatures). We immediately see that until the arrival of transformers, the trend of the landmark models was to increase their linearity, rather than to decrease it. On a more fine-grained level, we note that pure convolution architectures such as Alexnet (2012) and VGGs (2014) exhibit a very low spread of the affinity score values. This trend changes with the arrival of the inception module first used in Googlenet (2014): the latter includes activation functions that extend the range of the non-linearity on both ends of the spectrum. Importantly, we can see that the trend toward increasing the maximum and average non-linearity of the neural networks has continued for almost the whole decade. Even more surprisingly, EfficientNet models (2019), trained through neural architecture search, have strong negative skewness toward higher linearity, although they were state-of-the-art in their time. The second surprising finding comes with the arrival of ViTs (2020): they break the trend and leverage the non-linearity of their hidden activation functions becoming more or more non-linear with the varying size of the patches (see \cref{ax:raw_signatures} for a more detailed comparison with raw signatures). This trend remains valid also for Swin transformers (2021), although introducing the computer vision priors into them makes their non-linearity signature look more similar to pure convolutional networks from the early 2010s, such as Alexnet and VGGs. Finally, we observe that the non-linearity signature of a modern Convnext architecture (2022), designed as a convnet using best practices of Swin transformers, further confirms this. \\
\noindent \textbf{Similarities between architectures \phantom{.}} In \cref{fig:timeline} (B) and (C) we highlight the differences between the DNNs discussed earlier. For this, we first calculate the pairwise dynamic time warping (DTW) distances \cite{sakoe1978dynamic} between the non-linearity signatures of the representative models of each architecture. We plot the obtained result in \cref{fig:timeline} (B). From the first column of it, we can see that models proposed over the years were becoming more and more dissimilar to Alexnet, until making a ``U-turn'' with the introduction of Swin transformer and Convnext. Likewise, we note that ViT is highly dissimilar to all other models highlighting its uniqueness. Finally, we qualitatively confirm the findings from \cref{fig:timeline} (A) and (B) by plotting in \cref{fig:timeline} (C) the estimated kernel density estimates of the representatives of each architecture. We see that Alexnet has very similar non-linearity signature to Swin Base model, with the latter having heavier tails. This also holds for VGG16 and Convnext. One can note the distinct bi-modal structure of Resnet that we discuss more later. \\
\begin{figure}
    \centering
    \includegraphics[width=.9\linewidth]{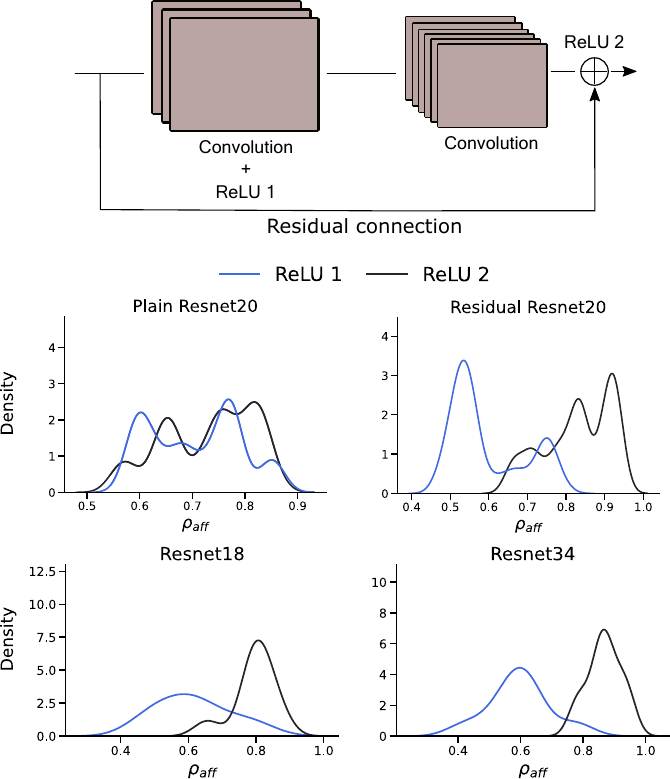}
    \caption{Comparing the same convnet with 20 layers when trained with (Residual Resnet20) and without (Plain Resnet20) residual connections (top row). Residual connections introduce a clear trend toward a bimodal distribution of affinity scores; the same effect is observed for Resnet18 and Resnet34 (bottom row).}
    \label{fig:resnets}
\end{figure}
\noindent \textbf{Closer look at accuracy/non-linearity trade-off \phantom{.}} Different families of vision models leverage different characteristics of their internal non-linearity to achieve better performance. To better understand this phenomenon, we now turn our attention to a more detailed analysis of the accuracy/non-linearity trade-off by looking for a statistic extracted from their non-linearity signatures that is the most predictive of their accuracy as measured by the $R^2$ score. Additionally, we also want to understand whether the non-linearity of DNNs can explain their performance better than the traditional characteristics such as the number of parameters, the number of giga floating point operations per second (GFLOPS), and the depth.
From the results presented in \cref{fig:correlations_accuracy}, we observe the following. First, the information extracted from the non-linearity signatures often correlates more with the final accuracy, than the usual DNN characteristics. This is the case for Residual networks (ResNets and DenseNets), ViTs, and vision models influenced by transformers (Post-ViT). Unsurprisingly, for models based on neural architecture search (NAS-based, i.e. EfficientNets and MNASNets) the number of parameters is the most informative metric as they are specifically designed to reach the highest accuracy with the increasing model size and compute. For Pre-residual pure convolutional models (Alexnet, VGGs, Googlenet, and Inception), the spread of the non-linearity explains the accuracy increase similarly to depth. Second, we observe that all models preceding ViTs were implicitly optimizing the spread of their affinity score values to achieve better performance. After the arrival of the transformers, the observed trend is to increase either the median or the minimum values of the non-linearity. This suggests a fundamental shift in the implicit bias that the transformers carry.  \\
\noindent \textbf{Role of residual connections \phantom{.}} From \cref{fig:timeline} (C), we note that Resnet is the first architecture that has a distinct bi-modal distribution of affinity scores. As this landmark architecture introduced for the first time the residual connections, we decide to explore what exact effect they have on the non-linearity signature of the same model trained with and without residual connection. \cref{fig:resnets} reveals vividly that residual connections make the distribution of the affinity scores bimodal with one such mode centered around highly linear activation functions. This confirms in a principled way that residual connections indeed tend to enable the learning of the identity function just as suggested in the seminal work that proposed them \cite{he2016deep}. 
Finally, we note that only ResNets exhibit this particular bi-modal distribution, thanks to the activation functions being exactly after the residual connections (illustrated in \cref{fig:resnets} (top)). However, we still observe rather high maximum affinity score in all subsequent architectures that use residual connections. \\
\noindent \textbf{Uniqueness of the affinity score \phantom{.}} No other metric extracted from the activation functions of the considered networks exhibits a strong consistent correlation with the non-linearity signature. To validate this claim, we compare in \cref{tab:correlations} the Pearson correlation between the non-linearity signature and several other metrics comparing the inputs and the outputs of the activation functions. %
We can see that for different models the non-linearity correlates with different metrics suggesting that it captures information that other metrics fail to capture consistently across all architectures. This becomes even more apparent when analyzing the individual correlation values (in \cref{ax:detailed_comp}). Overall, the proposed affinity score and the non-linearity signatures derived from it offer a unique perspective on the developments in ML. 
\begin{table}[!t]
    \centering
    \resizebox{\linewidth}{!}{%
    \newcommand{\pair}[2]{$#1$ {\color{gray}\scriptsize ($\pm #2$)}}
    \begin{tabular}{lccccc}
        Models & \textsc{cka} & \textsc{norm} & \textsc{sparsity} & \textsc{entropy} & $R^2$ \\
        \hline
        VGGs & \pair{0.0}{0.05}  & \pair{-0.67}{0.06} & \pair{-0.18}{0.03} & \pair{\textbf{-0.90}}{0.04} & \pair{-0.21}{0.06} \\
        ResNets & \pair{0.53}{0.04} & \pair{-0.41}{0.19} & \pair{\textbf{-0.68}}{0.02} & \pair{-0.38}{0.12} & \pair{-0.48}{0.24} \\
        DenseNets & \pair{0.88}{0.02} & \pair{-0.76}{0.02} & \pair{\textbf{-0.89}}{0.02} & \pair{-0.66}{0.03} & \pair{0.85}{0.04} \\
        MNASNets & \pair{\textbf{0.67}}{0.11} & \pair{-0.54}{0.14} & \pair{-0.63}{0.07} & \pair{-0.55}{0.16} & \pair{0.45}{0.17} \\
        EffNets & \pair{\textbf{0.42}}{0.10} & \pair{-0.16}{0.22} & \pair{-0.17}{0.23} & \pair{-0.16}{0.14} & \pair{0.21}{0.12} \\
        ViTs & \pair{-0.22}{0.40} & \pair{\textbf{-0.67}}{0.20} & \pair{-0.09}{0.56} & \pair{0.17}{0.25} & \pair{-0.10}{0.34} \\
        Swins & \pair{-0.15}{0.13} & \pair{\textbf{-0.53}}{0.10} & \pair{-0.26}{0.17} & \pair{0.06}{0.35} & \pair{-0.13}{0.13} \\
        Convnexts & \pair{0.69}{0.08} & \pair{0.21}{0.15} & \pair{0.23}{0.16} & \pair{0.02}{0.09} & \pair{\textbf{0.79}}{0.05} \\
        \hline
        Average & \pair{0.33}{0.45} & \pair{\textbf{-0.44}}{0.34} & \pair{-0.32}{0.42} & \pair{-0.31}{0.39} & \pair{0.14}{0.49} \\
    \end{tabular}

    }
    \caption{Pearson correlations between the non-linearity signature and other metrics, for all the architectures evaluated in this study. The highest absolute value in each group is reported in \textbf{bold}.}
    \label{tab:correlations}
    \end{table}

%% file: sec/5_discussion.tex
\section{Discussions}

We proposed the affinity score, the first sound approach to measure non-linearity of activation functions in neural networks, and defined their non-linearity signature based on it. We then showed what impacts the non-linearity of activation functions, both in isolation and as part of a randomly initialized DNN. We further used non-linearity signatures to provide a meaningful overview of the evolution of neural architectures proposed over the last decade with clear interpretable patterns. We showed the distinct features of the different landmark architectures and highlighted that the affinity score is unique in identifying them as it does not correlate with any other reasonable candidate metric. \\
\textbf{Future work and applications\phantom{.}} Our work can be applied to study non-linearity of large-scale foundation models to better understand the effect of different architectural and training choices. As the computation of the affinity score is fully differentiable, one future application can be to control the bias induced by each activation function through a regularization of the affinity score. Deviations between signatures of the same architecture can be used to monitor or identify the peculiarities of different training strategies since they lead to different non-linearity signatures (see \cref{sec:appendix_ssl}).

\section*{Acknowledgements}
This work was supported by the EU Horizon project ELIAS (No. 101120237) and by the French National
Research Agency (ANR) under the PEPR IA FOUNDRY (ANR-23-PEIA-0003), the LIMPID (ANR-20-CE23-0028) and the FAR-SEE (ANR-24-CE23-0921) projects. We also thank Florence d'Alché-Buc, Rémi Flamary and Devis Tuia for providing feedback on early versions of the manuscript.

%% file: sec/X_suppl.tex
\clearpage

\appendix
\section{Main theoretical results} \label{appendix:proofs}
\subsection{Proofs}
\label{subsec:proofs}
In this section, we provide proofs of the main theoretical results from the paper. 

\paragraph{Corollary 3.2.} Without loss of generality, let $X,Y\in \Probs_2(\R^d)$ be centered, and such that $Y=TX$, where $T$ is a positive semi-definite linear transformation. Then, $T$ is the OT map from $X$ to $Y$.
\begin{proof}
We first prove that we can consider centered distributions without loss of generality. To this end, 
we note that
\begin{equation}\label{eq:wasserstein_split}
\begin{aligned}
    W_2^2(X,Y) = W_2^2(X-\E[X], &Y-\E[Y])\\
    &+\|\E[X]-\E[Y]\|^2,
\end{aligned}
\end{equation}
implying that splitting the $2$-Wasserstein distance into two independent terms concerning the $L^2$ distance between the means and the $2$-Wasserstein distance between the centered measures. 

Furthermore, if we have an OT map $T'$ between $X-\E[X]$ and $Y-\E[Y]$, then
\begin{equation}\label{eq:linear_to_affine}
    T(x) = T'(x-\E[X]) + \E[Y],
\end{equation}
is the OT map between $X$ and $Y$.

To prove the statement of the Corollary, we now need to apply Theorem~\ref{thm:OT_maps} to the convex $\phi(x)= x^T T x$, where $T$ is positive semi-definite.
\end{proof}

\paragraph{Theorem 3.3.} Let $X,Y\in \Probs_2(\R^d)$ be centered and $Y = TX$ for a positive definite matrix $T$. Let  $N_X \sim \N(\mu(X),\Sigma(X))$ and $N_Y \sim \N(\mu(Y),\Sigma(Y))$ be their normal approximations where $\mu$ and $\Sigma$ denote mean and covariance, respectively. Then, $W_2(N_X,N_Y) = W_2(X,Y)$
and $T = T_\aff$, where $T_\aff$ is the OT map between $N_X$ and $N_Y$ and can be calculated in closed-form
    \begin{equation}
        \begin{aligned}
            &T_\aff(x) = Ax + b,\\
            &A = \Sigma(Y)^\frac{1}{2}\left(\Sigma(Y)^\frac{1}{2}\Sigma(X)\Sigma(Y)^\frac{1}{2}\right)^{-\frac{1}{2}}\Sigma(Y)^\frac{1}{2},\\
            &b = \mu(Y) - A\mu(X).
        \end{aligned} 
    \end{equation}
\begin{proof}
Corollary~\ref{cor:affine_ot_map} states that $T$ is an OT map, and 
\begin{equation*}
    \Sigma(T N_X) = T\Sigma(X)T = \Sigma(Y).
\end{equation*}
Therefore, $T N_X = N_Y$, and by Theorem~\ref{thm:OT_maps}, $T$ is the OT map between $N_X$ and $N_Y$. Finally, we compute
\begin{equation*}
    \begin{aligned}
    W^2_2(N_X,N_Y) &= \Tr[\Sigma(X)] + \Tr[T\Sigma(X)T]\\
    &- 2\Tr[T^\frac{1}{2}\Sigma(X) T^\frac{1}{2}]\\
    &= \argmin\limits_{T:T(X) = Y}\E_X[\|X-T(X)\|^2]\\
    &= W_2^2(X,Y).
    \end{aligned}
\end{equation*}
\end{proof}

\paragraph{Proposition 3.5.} Let $X,Y\in \Probs_2(\R^d)$ and $N_X, N_Y$ be their normal approximations. Then, 
\begin{enumerate}
    \item $
    \left| W_2(N_X, N_Y) - W_2(X,Y)\right| \leq \frac{2\Tr\left[\left(\Sigma(X)\Sigma(Y)\right)^\frac{1}{2}\right]}{\sqrt{\Tr[\Sigma(X)] + \Tr[\Sigma(Y)]}}.$
    \item For $T_\aff$ as in (\ref{eq:affine_map}), $W_2(T_\aff X,Y) \leq \sqrt{2}\Tr\left[\Sigma(Y)\right]^\frac{1}{2}.$ 
\end{enumerate}
\begin{proof}
By Theorem~\ref{thm:gaussians_lower_bound_2was}, we have $W_2(N_X,N_Y) \leq W_2(X,Y)$. On the other hand,
\begin{equation*}
\begin{aligned}
    W^2_2(X,Y) &= \min\limits_{\gamma \in \ADM(X,Y)} \int_{\R^d \times \R^d} \|x-y\|^2 d\gamma(x,y)\\
    &\leq \int_{\R^d \times \R^d}\left(\|x\|^2 + \|y\|^2\right) d\gamma(x,y)\\
    &= \Tr[\Sigma(X)] + \Tr[\Sigma(Y)].
\end{aligned}
\end{equation*}
Combining the above inequalities, we get
\begin{equation*}
\begin{aligned}
    &\left|W_2(N_X,N_Y) - W_2(X,Y)\right|\\
    &\leq \left| \sqrt{\Tr[\Sigma(X)] + \Tr[\Sigma(Y)]} - W_2(N_X,N_Y)\right|.
\end{aligned}
\end{equation*}
Let $a = \Tr[\Sigma(X)] + \Tr[\Sigma(Y)]$, and so $W^2_2(N_X,N_Y) = a - b$, where $b = 2\Tr\left[\left(\Sigma(X)\Sigma(Y)\right)^\frac{1}{2}\right]$. Then the RHS of can be written as
\begin{equation*}
    \left|\sqrt{a} - \sqrt{a-b}\right| = \frac{|a - (a-b)|}{\sqrt{a} + \sqrt{a-b}} \leq \frac{b}{\sqrt{a}},
\end{equation*}
where the inequality follows from positivity of $W_2(N_X,N_Y)=\sqrt{a-b}$. Letting $X = T_\aff X$ in the obtained bound gives 2).
\end{proof}

\subsection{Analytical expression in 1D for ReLU}
\label{subsec:1d_relu}
Let $X\sim \mathcal{U}[b,a]$, $b<0$ and $a>0$. Furthermore, let $f:x \mapsto \mathrm{ReLU}(x) = x\chi(x\geq 0)$ and $Y=f(X)$. 

We are interested in whether the affinity score
\begin{equation}
\label{eq:affine_transport}
    \rho_{\aff}(X,Y) = 1 - \frac{W_2(T_{\aff}(X), Y)}{\sqrt{2\Tr \left[\Sigma(Y)\right]}}
\end{equation}
is symmetric wrt. $a$ and $b$. Here $W_2$ is the $2$-Wasserstein distance between the laws of two random variables and $T_{\aff}$ is the affine transport map between $X$ and $Y$ given by
\begin{equation}
    \begin{aligned}
        T_{\aff}(x) &= A_{\aff} x+ b_{\aff}, \\
        A_{\aff} &= \Sigma(Y)^\frac{1}{2}\left(
        \Sigma(Y)^\frac{1}{2}\Sigma(X)\Sigma(Y)^\frac{1}{2}
        \right)^{-\frac{1}{2}}\Sigma(Y)^\frac{1}{2}, \\
        b_{\aff} &= \mu(Y) - A_{\aff} \mu(X).
    \end{aligned}
\end{equation}

\textbf{Source Mean and Variance.}
Recall the formulae for mean and variance for a uniform distribution
$$
\mu(X) = \frac{a+b}{2}, \quad \Sigma(X)= \frac{(a-b)^2}{12}
$$

\textbf{Target Mean and Variance.}
This time we are forced to compute a bit. Let's tart with the mean
\begin{equation}
    \begin{aligned}
        \mu(Y) &= \frac{1}{a-b}\int_b^af(x)dx\\
        &= \frac{1}{a-b}\left(
        \int_b^0 0 dx + \int_0^a x dx
        \right) \\
        &= \frac{a^2}{2(a-b)}.
    \end{aligned}
\end{equation}

Moving on to the variance
\begin{equation}
\label{eq:var_y}
    \begin{aligned}
        \Sigma(Y) &= \frac{1}{a-b}\int_b^a\left(f(x) - \mu(Y)\right)^2dx \\
        &= \frac{1}{a-b}\left(\int_b^0 \mu(Y)^2 dx + \int_0^a\left(x-\mu(Y)\right)^2dx\right)\\
        &= \frac{1}{a-b} \left(
        (a-b)\mu(Y)^2 - a^2\mu(Y) + \frac{1}{3}a^3
        \right) \\
        &= \left(\frac{2a}{3}-\mu(Y)\right)\mu(Y) \\
        &= \frac{a^3(a-4b)}{12(a-b)^2}
    \end{aligned}
\end{equation}

\textbf{Affine transport map.}
Substituting the computed statistics into \eqref{eq:affine_transport} and abusing their scalar nature, we get
\begin{equation}
\begin{aligned}
    A_{\aff} &= \frac{\Sigma(Y)^\frac{1}{2}}{\Sigma(X)^\frac{1}{2}}\\
    &= \frac{\sqrt{a^3(a-4b)}}{(a-b)^2}, \\
    \quad b_{\aff} &= \mu(Y) - A_{\aff}\mu(X), \\
    &= \frac{a}{2(a-b)}\left(a - \left(\frac{a+b}{a-b}\right)\left(\sqrt{a(a-4b)}\right)\right)
\end{aligned}
\end{equation}

\textbf{$2$-Wasserstein Distance.} Recall that the $2$-Wasserstein distance between scalars is simply a sorting problem: sort the source and target and match the elements with similar indices. Luckily in our case, both $T_{\aff}$ and $f=ReLU$ preserve order as increasing functions, and hence
\begin{equation}
\label{eq:2-wasserstein-taff-relu}
    \begin{aligned}
        W_2^2\left(T_{\aff}(X), Y\right) &= W_2^2\left(T_{\aff}(X), ReLU(X)\right) \\
        &= \frac{1}{a-b}\int_b^a\left(T_{\aff}(x) - ReLU(x)\right)^2 dx \\
    \end{aligned}
\end{equation}

Before continuing the computation, remember that due to affine transport, $\mu(T(X)) = \mu(Y)$ and $\Sigma(T(X))=\Sigma(Y)$. Therefore
\begin{equation}
    \begin{aligned}
        \Sigma(T_{\aff}(X)) &= \Sigma(Y) \\
        \Rightarrow \Expect\left[T_{\aff}(X)^2\right] - \Expect\left[T_{\aff}(X)\right]^2 &= \Expect\left[Y^2\right] - \Expect\left[Y\right]^2 \\
        \Rightarrow \Expect\left[T_{\aff}(X)^2\right] &= \Expect\left[Y^2\right].
    \end{aligned}
\end{equation}

Using this, we can continue the computation in \eqref{eq:2-wasserstein-taff-relu}
\begin{align*}
        & W_2^2(T_{\aff}(X), Y) \\
        =& \frac{2}{a-b}\left(\int_b^a ReLU^2(x)dx - \int_b^a ReLU(x)T_{\aff}(x)dx\right) \\
        =& \frac{2}{a-b}\left(\int_0^a x^2dx - \int_0^a (A_{\aff}x^2 + b_{\aff}x) dx\right) \\
        =& \frac{a^2}{3(a-b)}\left(2a\left(1 - A_{\aff}\right) - 3b_{\aff}\right) \\
        =& \frac{2}{3}\mu(Y)(2a(1-A_{\aff}) - 3b_{\aff})\\
        = & \frac{a^3}{6(a-b)^2}\left((a-4b) + \sqrt{a(a-4b)}\left(\frac{-a+3b}{a-b}\right)\right).
\end{align*}

\newpage
\section{Affinity scores of other popular activation functions}
\label{sec:appendix_activations}

\begin{figure*}[!t]
    \centering
    \includegraphics[width=.32\linewidth]{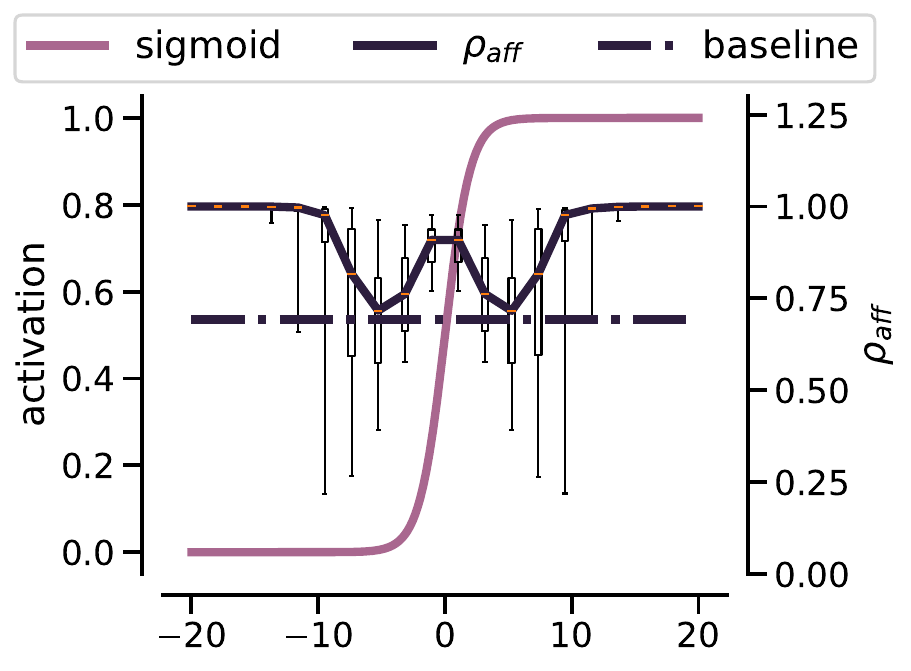}
    \includegraphics[width=.32\linewidth]{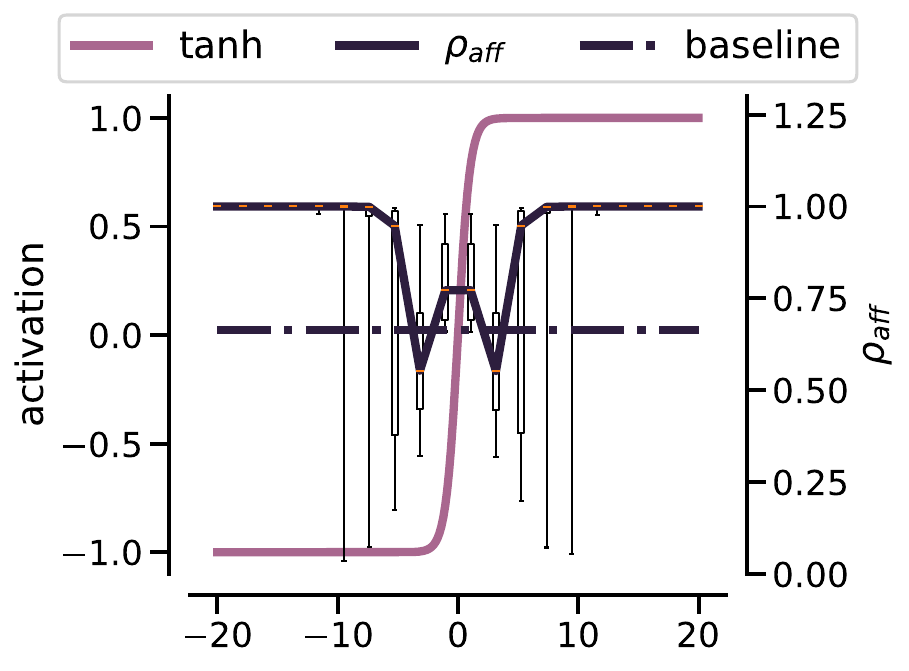}
     \includegraphics[width=.32\linewidth]{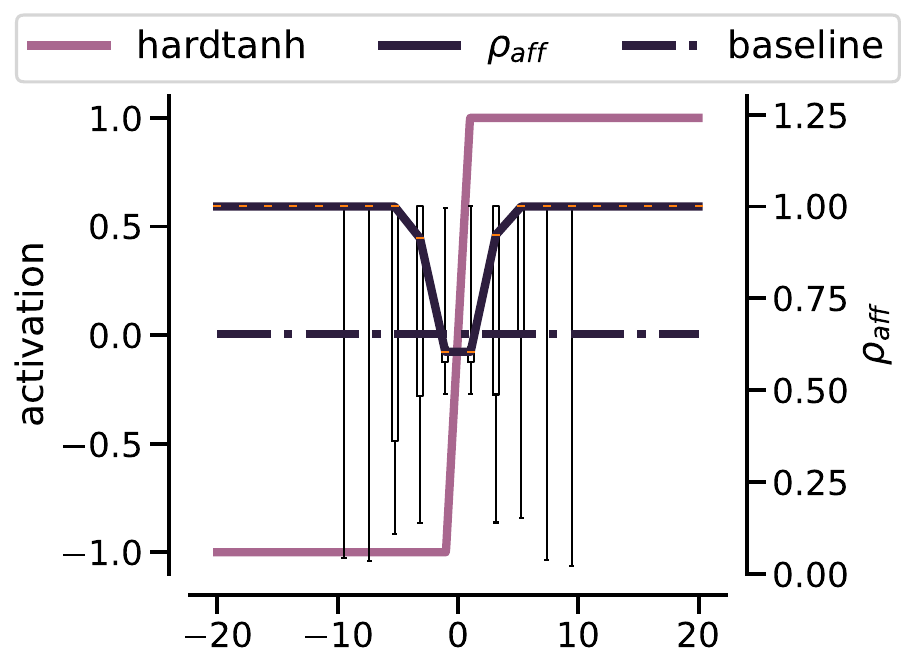}
    
    \includegraphics[width=.32\linewidth]{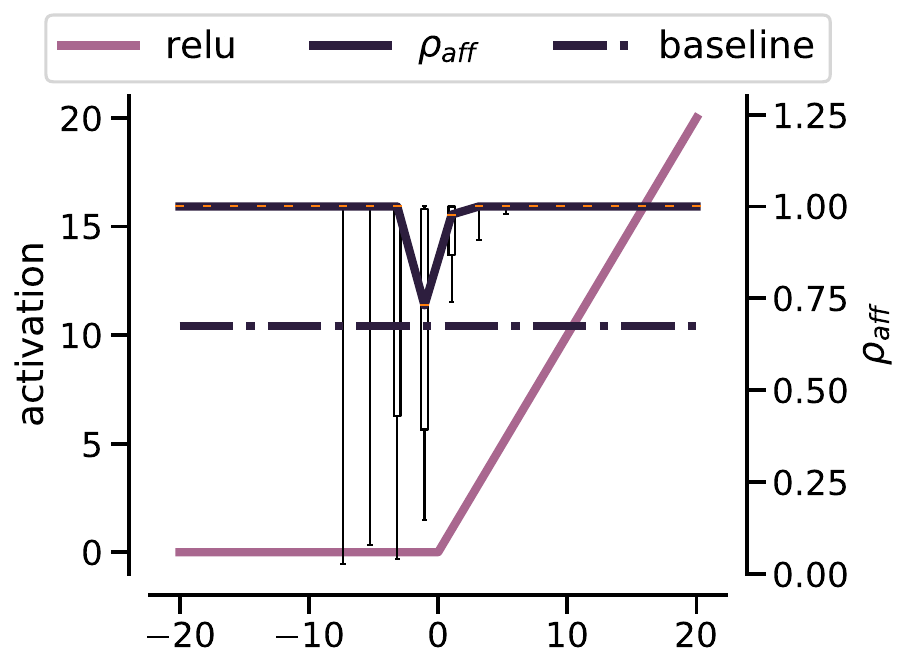}
    \includegraphics[width=.32\linewidth]{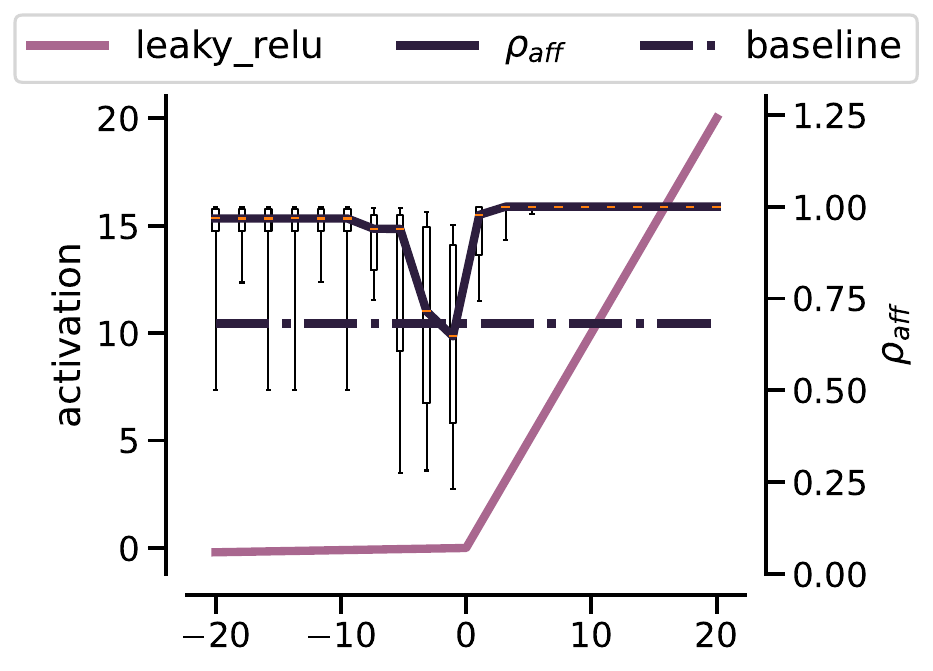}
     \includegraphics[width=.32\linewidth]{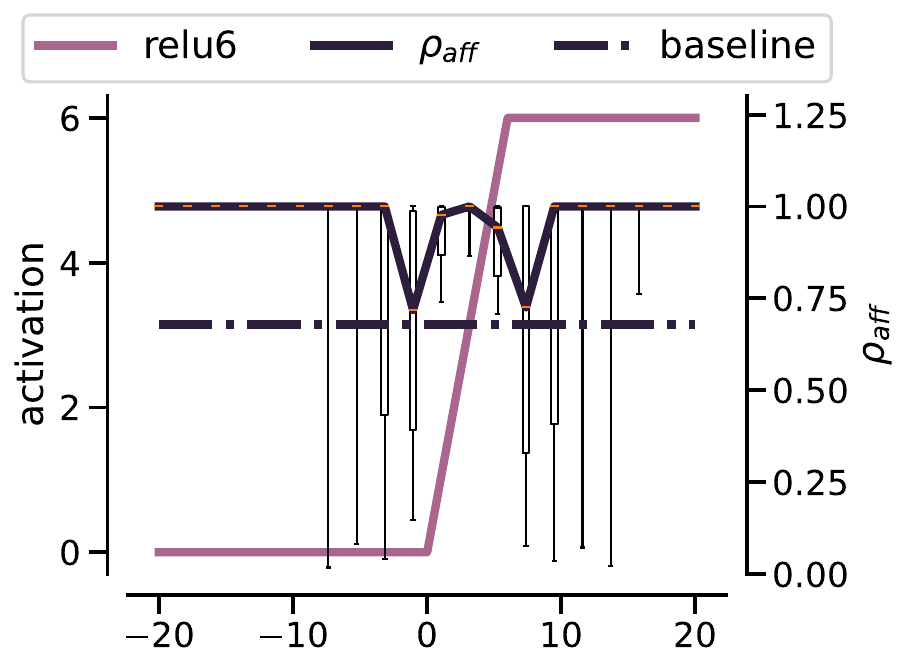} 
     
     \includegraphics[width=.32\linewidth]{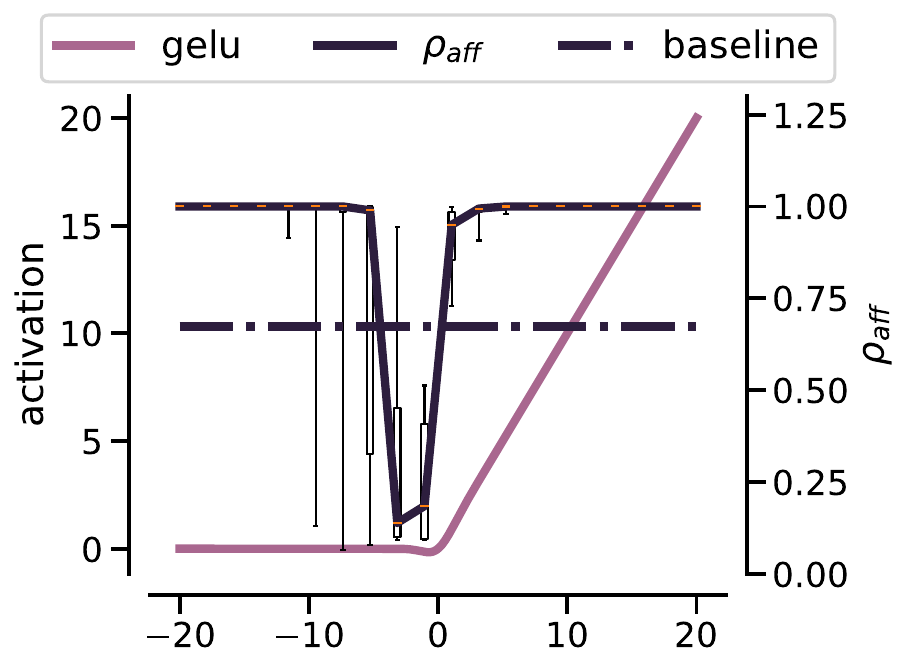}
     \includegraphics[width=.32\linewidth]{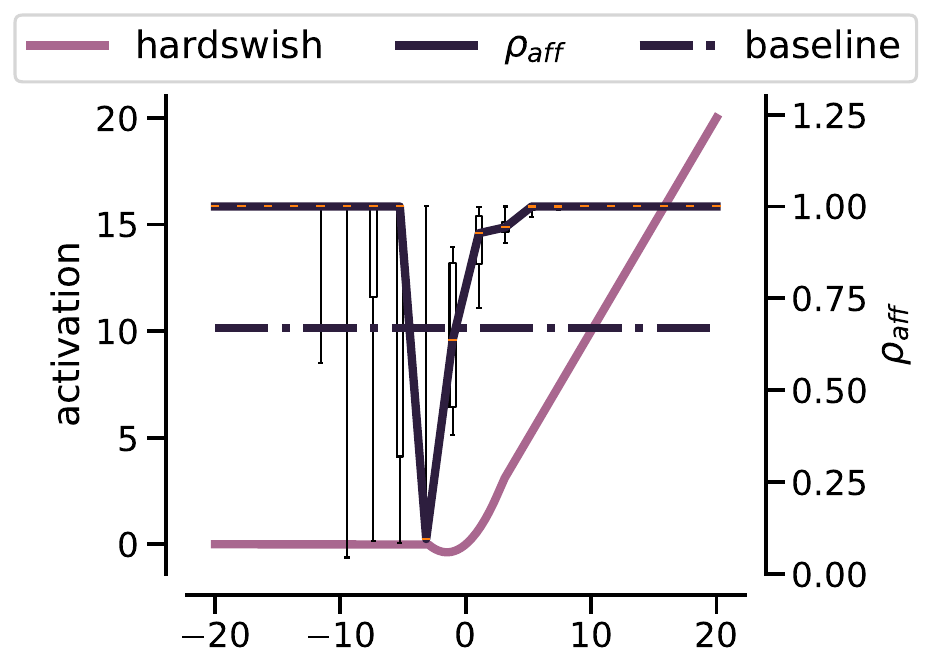}
     \includegraphics[width=.32\linewidth]{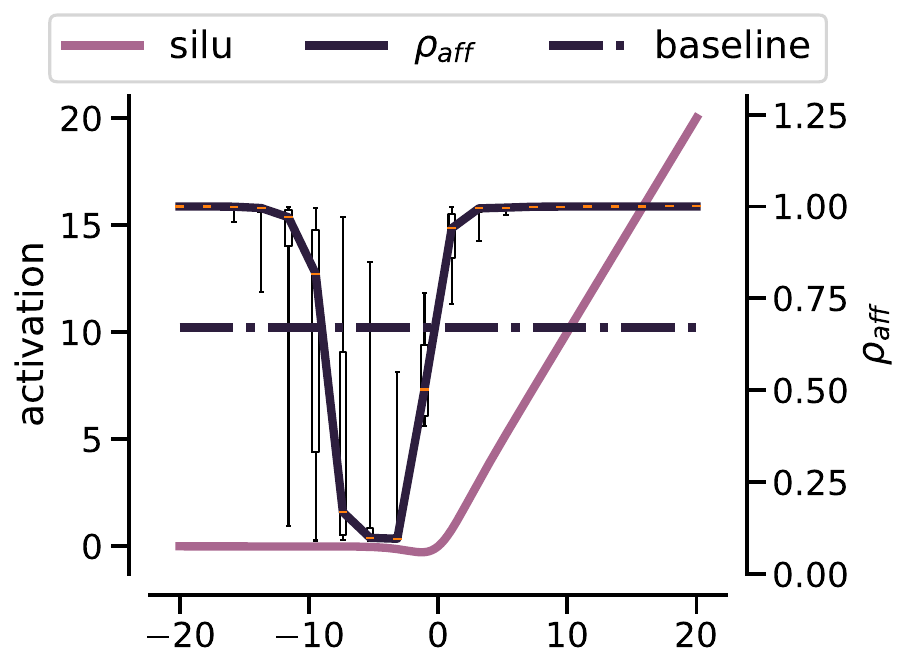}
    \caption{Median affinity scores of Sigmoid, ReLU, GELU, ReLU6, LeakyReLU with a default value of slope, Tanh, HardTanh, SiLU, and HardSwish obtained across random draws from Gaussian distribution with a sliding mean and varying stds used as their input. Whiskers of boxplots show the whole range of values obtained for each mean across all stds. The baseline value is the affinity score obtained for a sample covering the whole interval. The ranges and extreme values of each activation function over its subdomain are indicative of its non-linearity limits.}
    \label{fig:more_activations}
\end{figure*}

Many works aimed to improve the way how the non-linearity -- represented by activation functions -- can be defined in DNNs. As an example, a recent survey on the commonly used activation functions in deep neural networks \citep{survey_activations} identifies over 40 activation functions with first references to sigmoid dating back to the seminal paper \citep{rumelhart1986learning} published in late 80s. The fashion for activation functions used in deep neural networks evolved over the years in a substantial way, just as the neural architectures themselves. Saturating activations, such as sigmoid and hyperbolic tan, inspired by computational neuroscience were a number one choice up until the arrival of rectifier linear unit (ReLU) in 2010. After being the workhorse of many famous models over the years, the arrival of transformers popularized Gaussian Error Linear Unit (GELU) which is now commonly used in many large language models including GPTs.

We illustrate in \cref{fig:more_activations} the affinity scores obtained after a single pass of the data through the following activation functions: Sigmoid, ReLU \citep{pmlr-v15-glorot11a}, GELU \citep{hendrycks2016gaussian}, ReLU6 \citep{howard2017mobilenets}, LeakyReLU \citep{maas2013rectifier} with a default value of the slope, Tanh, HardTanh, SiLU \citep{elfwing2018sigmoid}, and HardSwish \citep{howard2019searching}. As the non-linearity of activation functions depends on the domain of their input, we fix 20 points in their domain equally spread in $[-20,20]$ interval. We use these points as means $\{m_i\}_{i=1}^{20}$ of Gaussian distributions from which we sample $1000$ points in $\mathbb{R}^{300}$ with standard deviation (std) $\sigma$ taking values in $[2, 1, 0.5, 0.25, 0.1, 0.01]$. Each sample denoted by $X_{m_i}^{\sigma_j}$ is then passed through the activation function $\text{act} \in \{\text{sigmoid}, \text{ReLU}, \text{GELU}\}$ to obtain $\rho_\text{aff}^{m_i,\sigma_j} := \rho_\text{aff}(X_{m_i}^{\sigma_j}, \text{act}(X_{m_i}^{\sigma_j}))$. Larger std values make it more likely to draw samples that are closer to the region where the studied activation functions become non-linear. We present the obtained results in Figure S2 where each of 20 boxplots showcases $\text{median}(\rho_\text{aff}^{m_i,\sigma_\cdot})$ values with 50\% confidence intervals and whiskers covering the whole range of obtained values across all $\sigma_j$. 

This plot allows us to derive several important conclusions. We observe that each activation function can be characterized by 1) the lowest values of its non-linearity obtained for some subdomain of the considered interval and 2) the width of the interval in which it maintains its non-linearity. We note that in terms of 1) both GELU and ReLU may attain affinity scores that are close to 0, which is not the case for Sigmoid. For 2), we observe that the non-linearity of Sigmoid and GELU is maintained in a wide range, while for ReLU it is rather narrow. We can also see a distinct pattern of more modern activation functions, such as SiLU and HardSwish having a stronger non-linearity pattern in large subdomains. We also note that despite having a shape similar to Sigmoid, Tanh may allow for much lower affinity scores. Finally, the variations of ReLU seem to have a very similar shape with LeakyReLU being on average more linear than ReLU and ReLU6.

\newpage
\section{Implementation details}\label{ax:impl_details}

\begin{figure}[!t]
    \centering
    \includegraphics[width=0.45\textwidth]{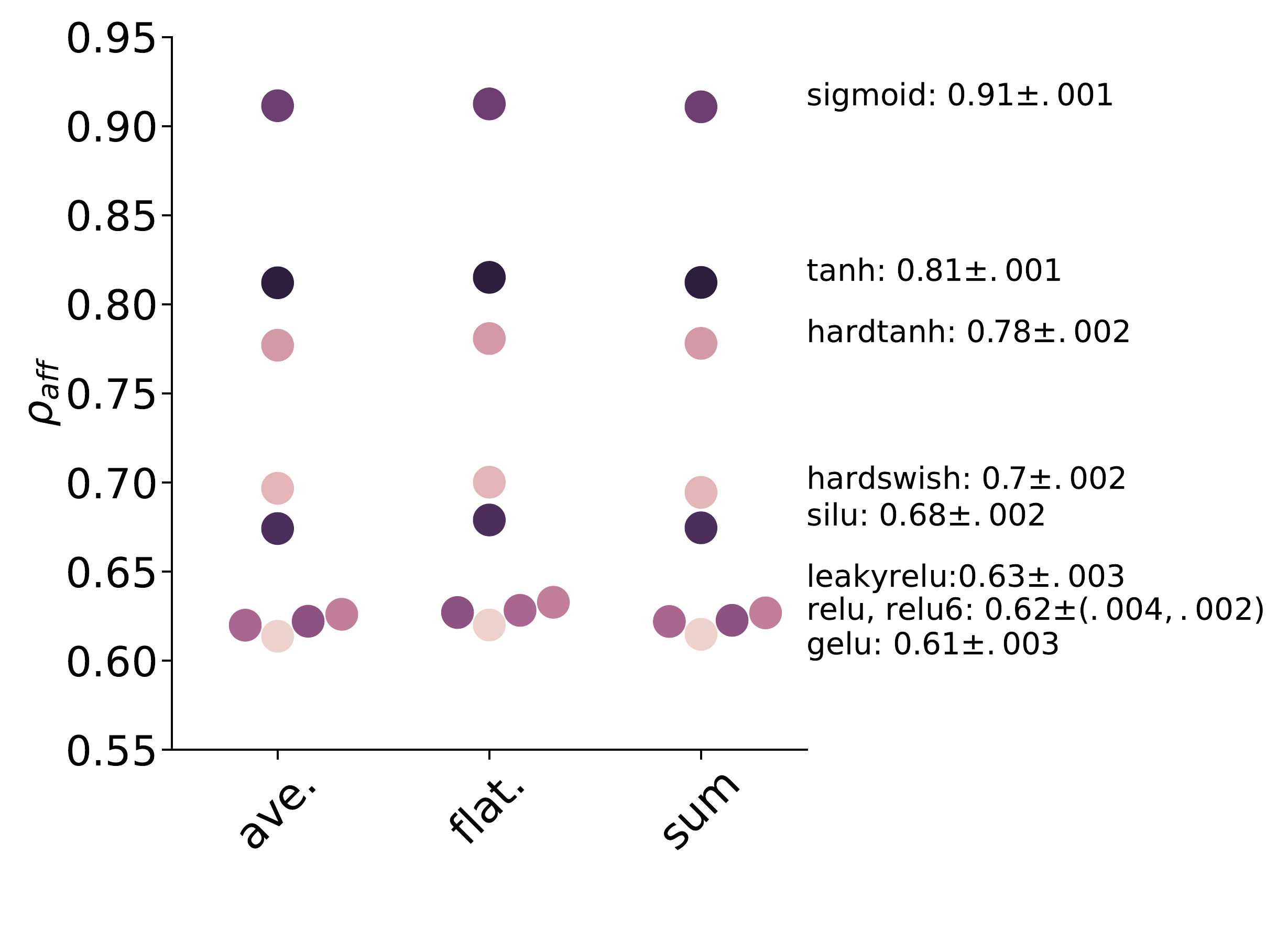}
    \includegraphics[width=0.45\textwidth]{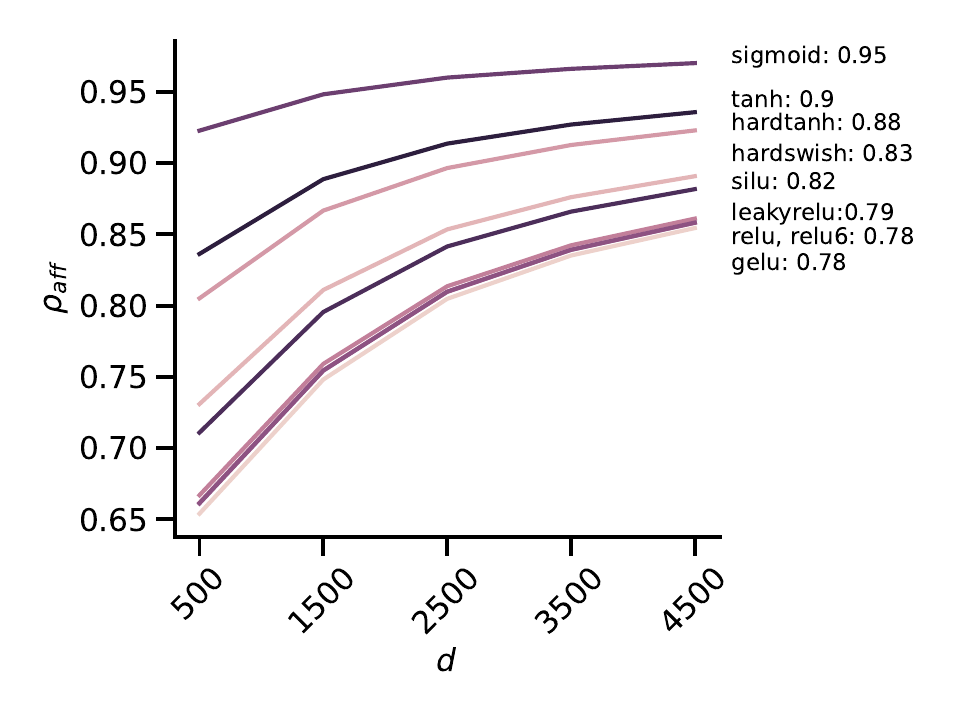}
     \includegraphics[width=0.45\textwidth]{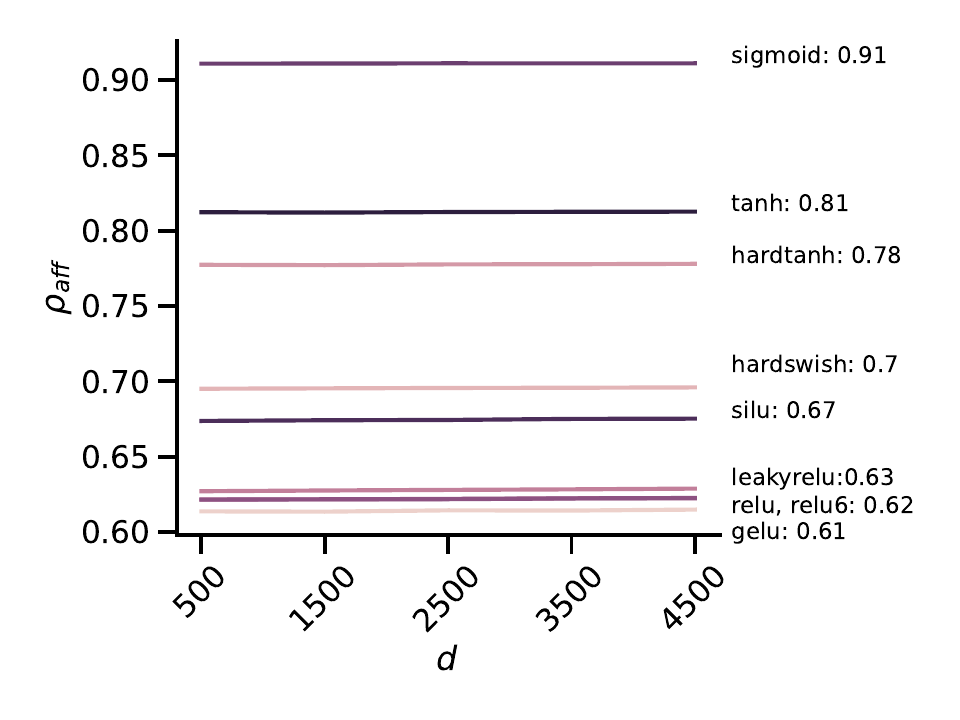}
    \caption{\textbf{(Top)} Affinity score is robust to the dimensionality reduction both when using averaging and summation over the spatial dimensions; \textbf{(Middle)} When $d>n$, sample covariance matrix estimation leads to a lack of robustness in the estimation of the affinity score; \textbf{(Bottom)} Shrinkage of the covariance matrix leads to constant values of the affinity scores with increasing $d$.} 
    \label{fig:shrink_pooling}
\end{figure}

\paragraph{Dimensionality reduction} Manipulating 4-order tensors is computationally prohibitive and thus we need to find an appropriate lossless function $r$ to facilitate this task. One possible choice for $r$ may be a vectorization operator that flattens each tensor into a vector. In practice, however, such flattening still leads to very high-dimensional data representations. In our work, we propose to use averaging over the spatial dimensions to get a suitable representation of the manipulated tensors. In \cref{fig:shrink_pooling} (top), we show that the affinity score is robust wrt such an averaging scheme and maintains the same values as its flattened counterpart. 

\paragraph{Computational considerations} The non-linearity signature requires calculating the affinity score over ``wide'' matrices. Indeed, after the reduction step is applied to a batch of $n$ tensors of size $h\times w \times c$, we end up with matrices of size $n\times c$ where $n$ may be much smaller than $c$. This is also the case when input tensors are 2D when the batch size is smaller than the dimensionality of the embedding space. To obtain a well-defined estimate of the covariance matrix in this case, we use a known tool from the statistics literature called Ledoit-Wolfe shrinkage \citep{ledoit}. In \cref{fig:shrink_pooling} (bottom), we show that shrinkage allows us to obtain a stable estimate of the affinity scores that remain constant in all regimes. 

\begin{figure}[!t]
    \centering
    \includegraphics[width=.49\linewidth]{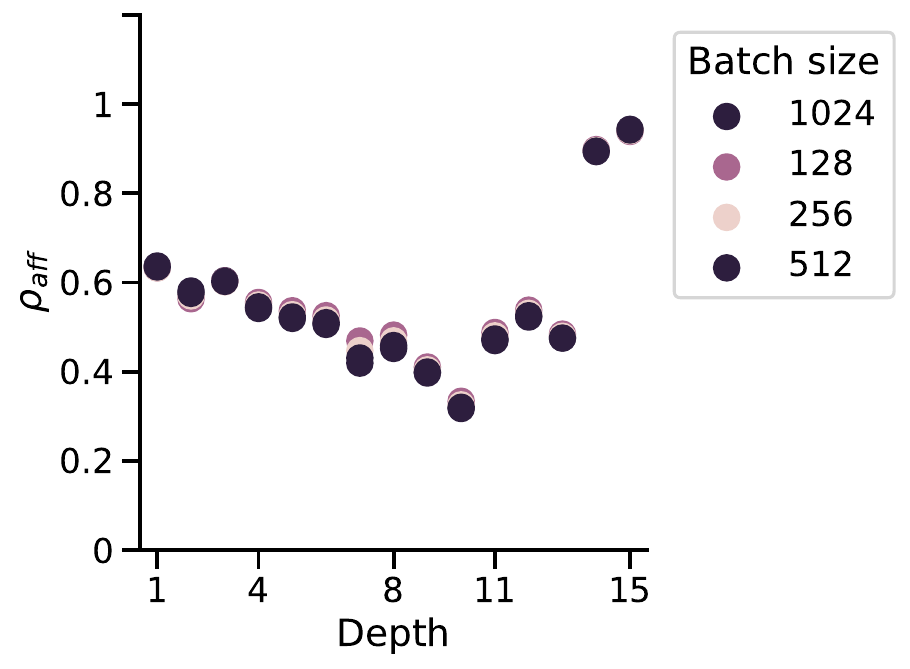}
    \includegraphics[width=.49\linewidth]{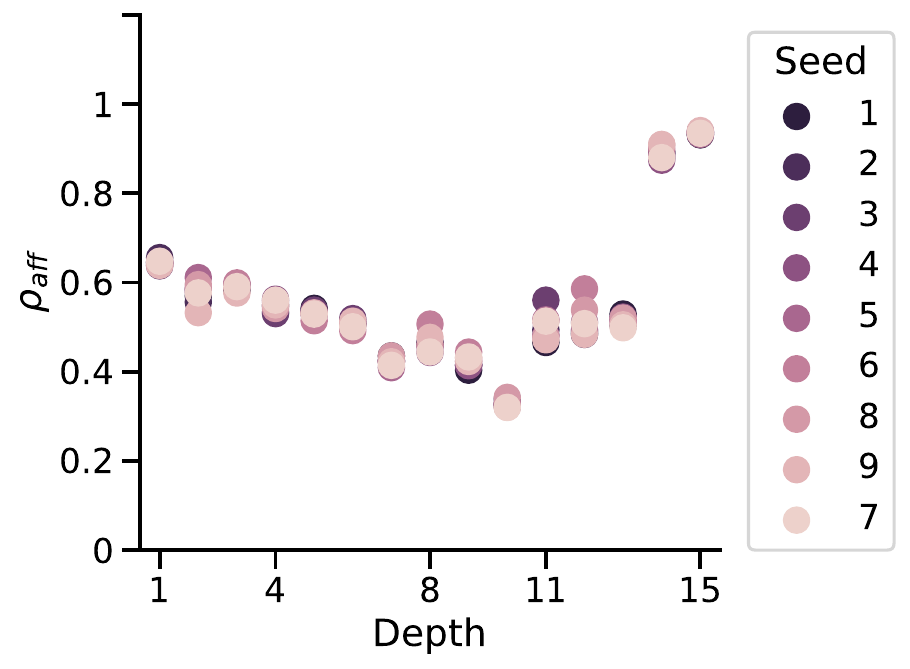}
    \caption{Non-linearity signature of VGG16 on CIFAR10 with a varying batch size (left) and when retrained from 9 different random seeds (right).}
    \label{fig:appendix_robustness}
\end{figure}

\paragraph{Robustness to batch size and different seeds} In this section, we highlight the robustness of the non-linearity signature with respect to the batch size and the random seed used for training. To this end, we concentrate on VGG16 architecture and CIFAR10 dataset to avoid costly Imagenet retraining. In \cref{fig:appendix_robustness}, we present the obtained result where the batch size was varied between 128 and 1024 with an increment of 128 (left plot) and when VGG16 model was retrained with seeds varying from 1 to 9 (right plot). The obtained results show that the affinity score is robust to these parameters suggesting that the obtained results are not subject to a strong stochasticity. 

\begin{figure}[t]
    \centering
    \includegraphics[width=.75\linewidth]{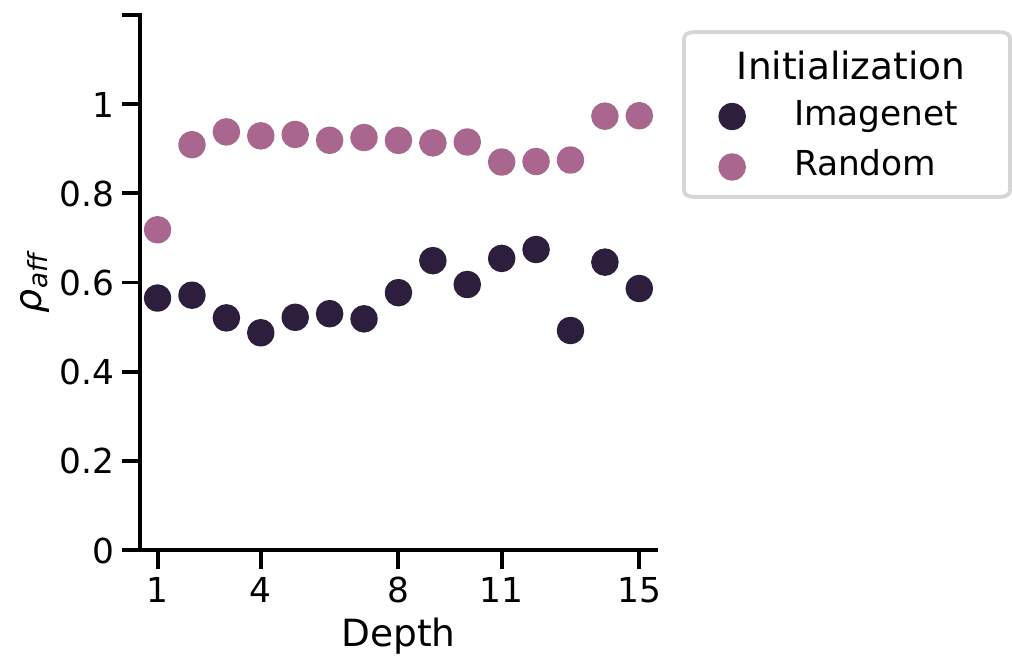}
    \caption{Non-linearity signatures of VGG16 on CIFAR10 in the beginning and end of training on Imagenet.}
    \label{fig:appendix_training_comparison}
\end{figure}

\paragraph{Impact of training} Finally, we also show how a non-linearity signature of a VGG16 model looks like at the beginning and in the end of training on Imagenet. We extract its non-linearity signature at initialization when making a feedforward pass over the whole CIFAR10 dataset and compare it to the non-linearity signature obtained in the end. In \cref{fig:appendix_training_comparison}, we can see that at initialization the network's non-linearity signature is increasing, reaching almost a perfectly linear pattern in the last layers. Training the network enhances the non-linearity in a non-monotone way. Importantly, it also highlights that the non-linearity signature is capturing information from the training process.

\newpage
\begin{figure*}[tb]
    \centering
     \includegraphics[width=.3\linewidth]{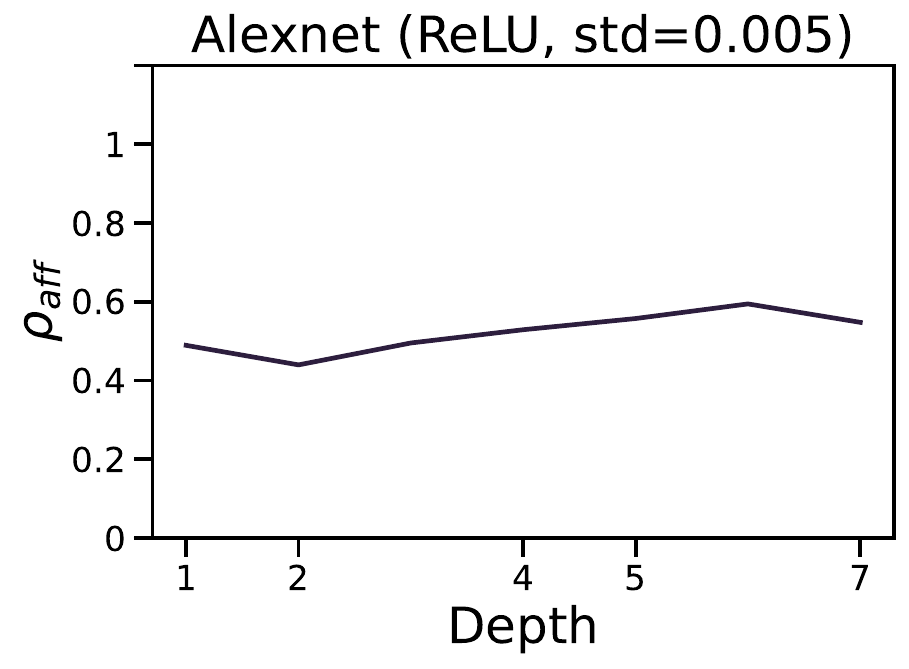}\hfill
     \includegraphics[width=.3\linewidth]{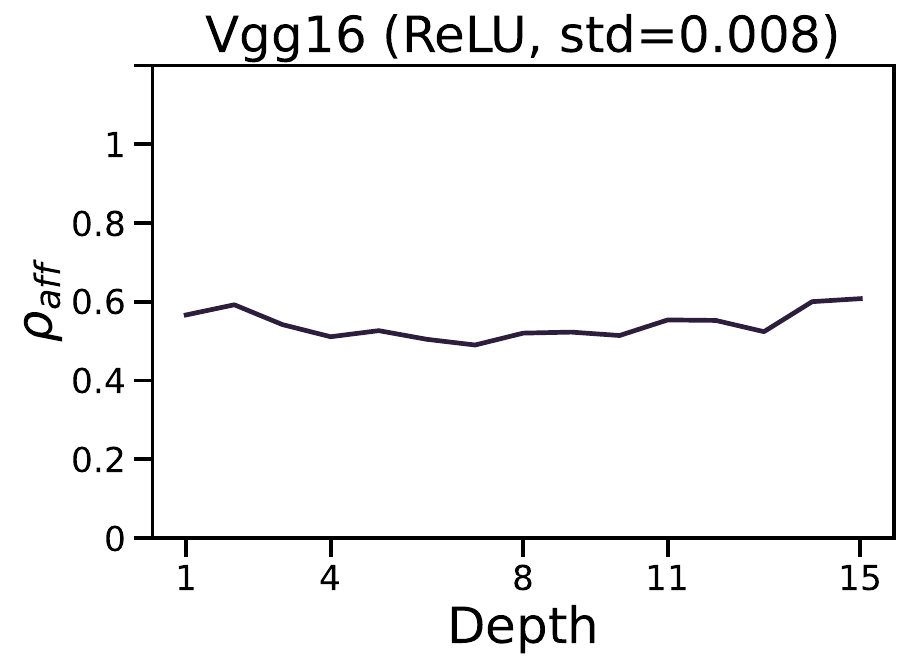} \hfill
     \includegraphics[width=.305\linewidth]{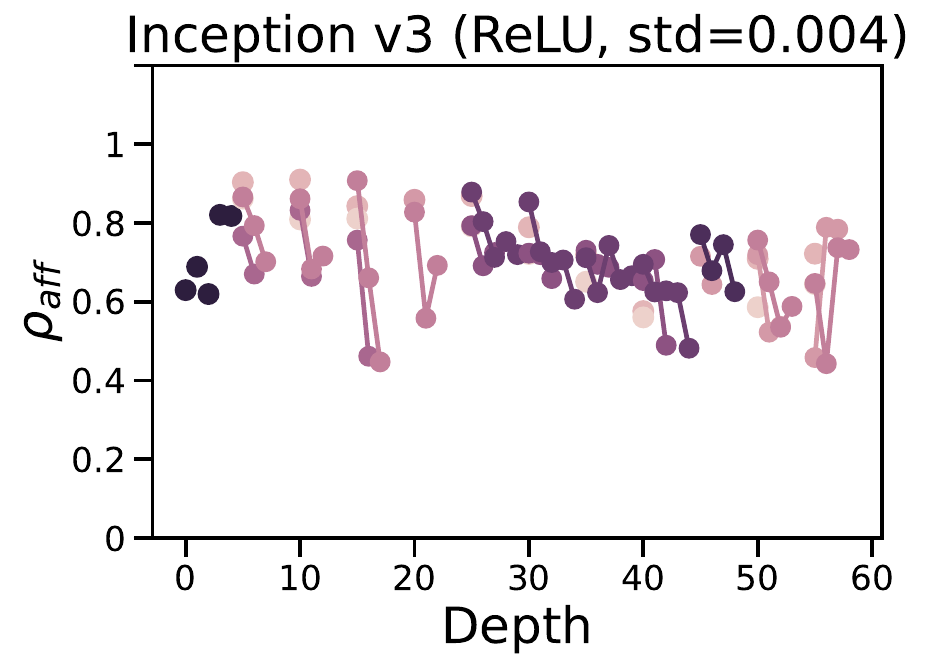}
     
     \includegraphics[width=.3\linewidth]{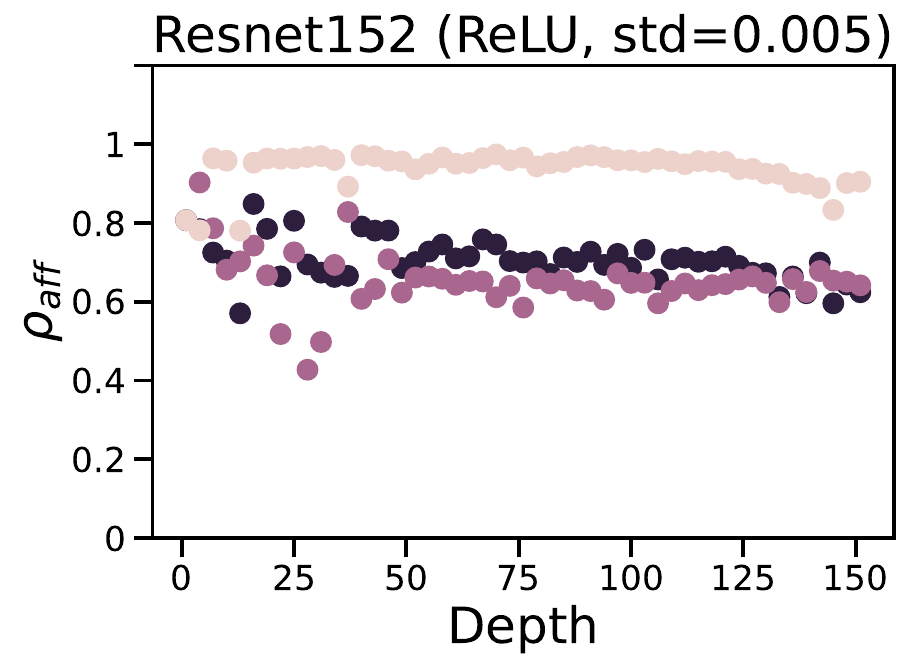} \hfill
     \includegraphics[width=.31\linewidth]{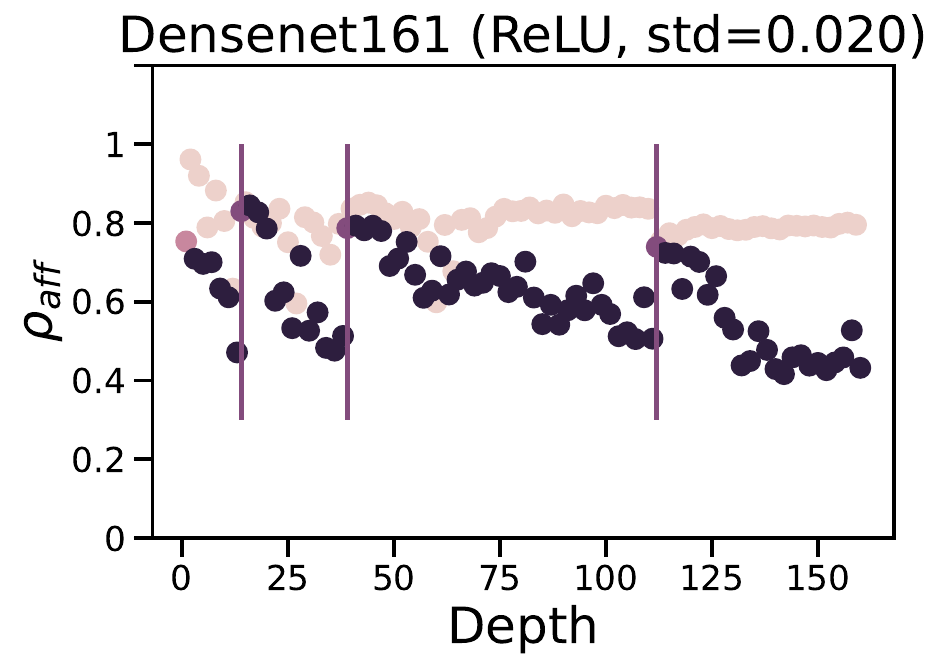}\hfill
     \includegraphics[width=.3\linewidth]{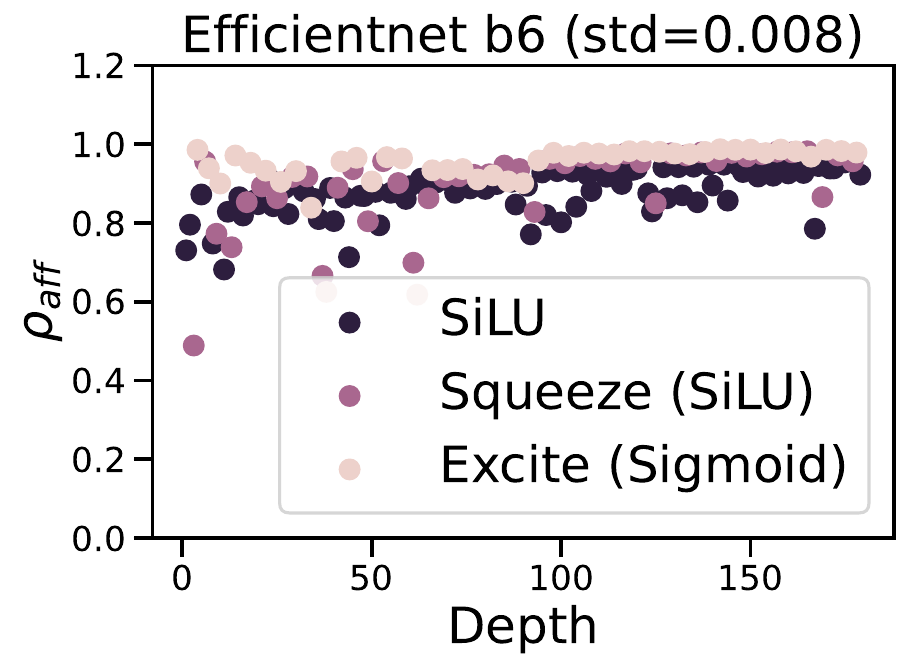}

     \includegraphics[width=.315\linewidth]{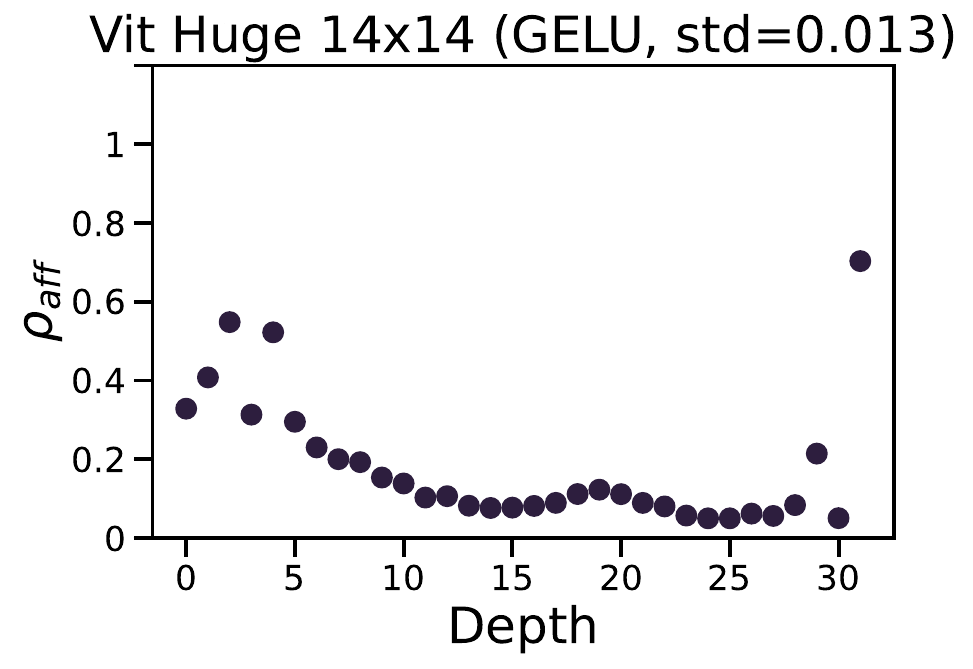}\hfill 
     \includegraphics[width=.3\linewidth]{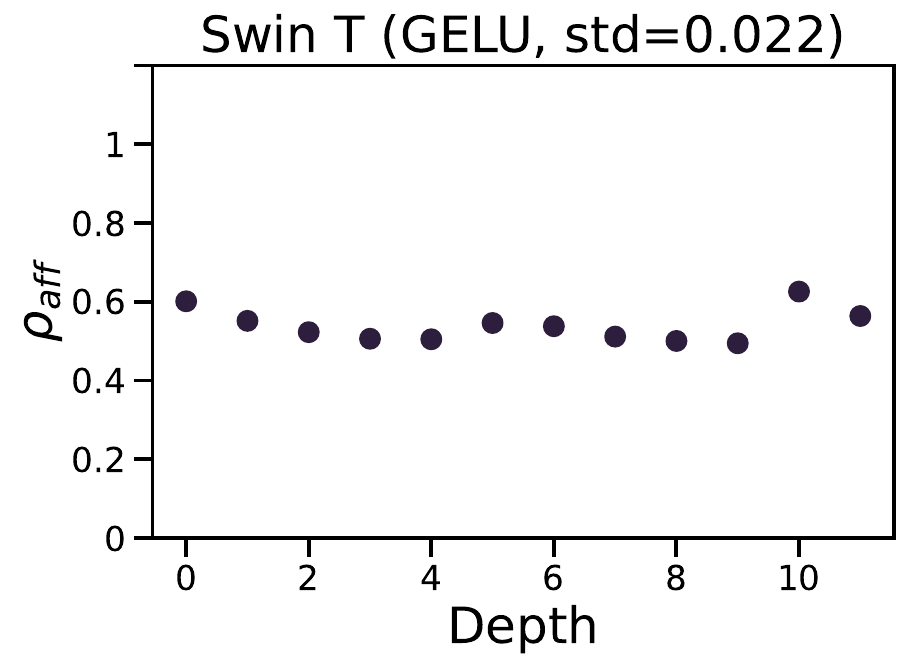}\hfill
     \includegraphics[width=.3\linewidth]{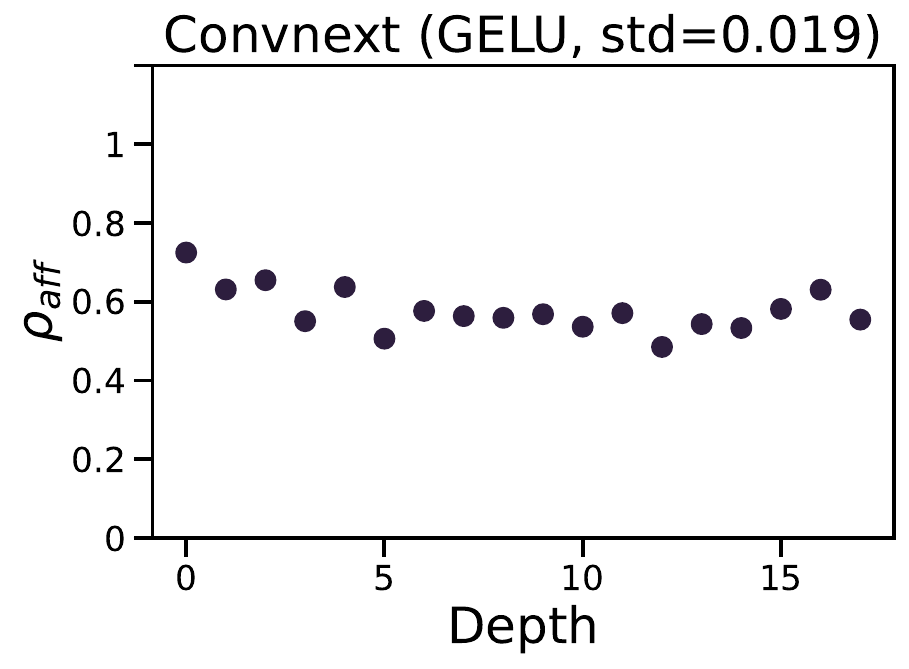}
    \caption{Raw non-linearity signatures of popular DNN architectures, plotted as affinity scores over the depth throughout the network.}
    \label{fig:raw_signatures}
\end{figure*}

\begin{figure*}[t]
    \centering
    \includegraphics[width=.3\linewidth]{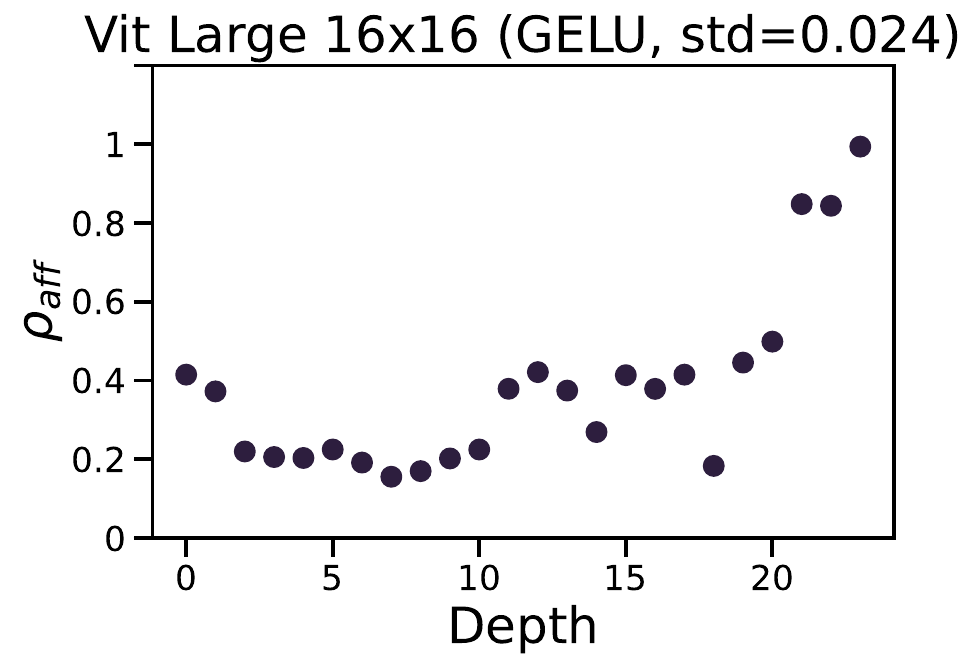}\hfill
    \includegraphics[width=.3\linewidth]{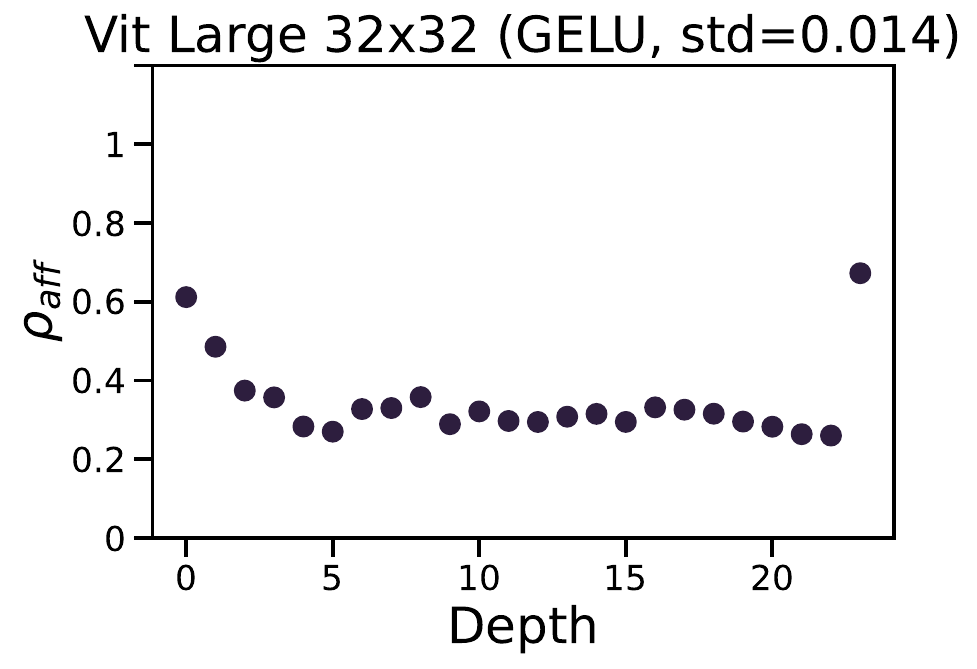}\hfill
    \includegraphics[width=.3\linewidth]{figures/signatures/vit_aff_scores_image.pdf}
    \caption{ViTs: Large ViT with 16x16 and 32x32 patch sizes and Huge ViT.}
    \label{fig:vit_signatures}
\end{figure*}

\section{Raw signatures}\label{ax:raw_signatures}
In \cref{fig:raw_signatures}, we portray the raw non-linearity signatures of several representative networks studied in the main paper. We use different color codes for distinct activation functions appearing repeatedly in the considered architecture (for instance, every first ReLU in a residual block of a Resnet). We also indicate the mean standard deviation of the affinity scores over batches in the title.

We see that the non-linearities across ReLU activations in all of Alexnet's 8 layers remain stable. Its successor, VGG network, reveals tiny, yet observable, variations in the non-linearity propagation with increasing depth and, slightly lower overall non-linearity values. We attribute this to the decreased size of the convolutional filters (3x3 vs. 7x7). The Googlenet architecture was the first model to consider learning features at different scales in parallel within the so-called inception modules. This add more variability as affinity scores of activation in Googlenet vary between 0.6 and 0.9. Despite being almost 20 times smaller than VGG16, the accuracy of Googlenet on Imagenet remains comparable, suggesting that increasing and varying the linearity is a way to have high accuracy with a limited computational complexity compared to predecessors. This finding is further confirmed with Inception v3 that pushed the spread of the affinity score toward being more linear in some hidden layers. When comparing this behavior with Alexnet, we note just how far we are from it. Resnets achieve the same spread of values of the non-linearity but in a different, and arguably, simpler way. Indeed, the activation after the skip connection exhibits affinity scores close to 1, while the activations in the hidden layers remain much lower. Densenet, that connect each layer to all previous layers and not just to the one that precedes it, is slightly more non-linear than Resnet152, although the two bear a striking similarity: they both have an activation function that maintains the non-linearity low with increasing depth. Additionally, transition layers in Densenet act as linearizers and allow it to reset the non-linearity propagation in the network by reducing the feature map size. ViTs (Large with 16x16 and 32x32 patch sizes, and Huge with 14x14 patches) are all highly non-linear models to the degree yet unseen. Interestingly, as seen in \cref{fig:vit_signatures} the patch size affects the non-linearity propagation in a non-trivial way: for 16x16 size a model is more non-linear in the early layers, while gradually becoming more and more linear later, while 32x32 patch size leads to a plateau in the hidden layers of MLP blocks, with a steep change toward linearity only in the final layer. We hypothesize that attention modules in ViT act as a focusing lens and output the embeddings in the domain where the activation function is the most non-linear.

\begin{figure*}[t]
    \centering
    \includegraphics[width=.24\linewidth]{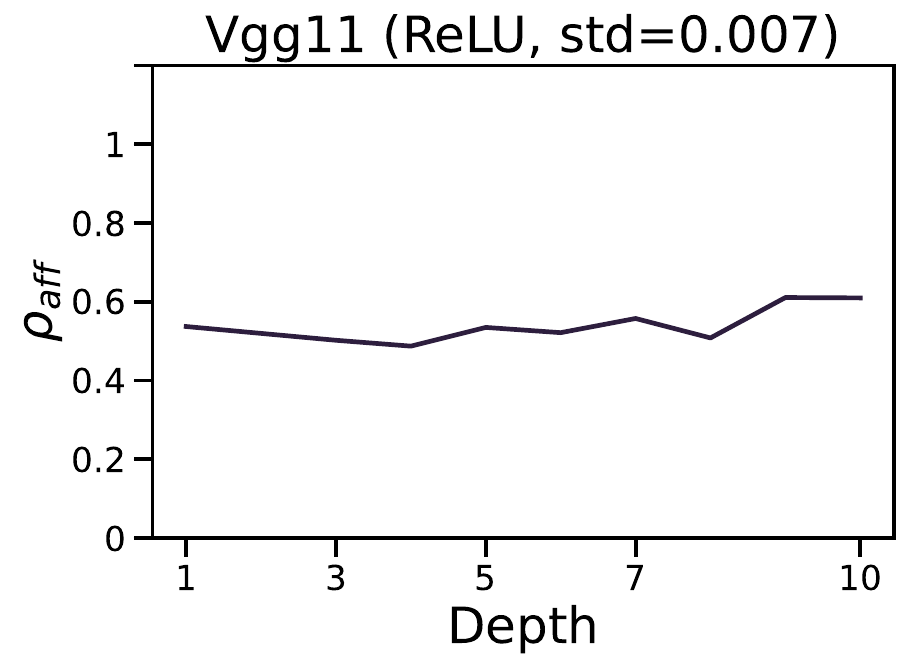}
    \includegraphics[width=.24\linewidth]{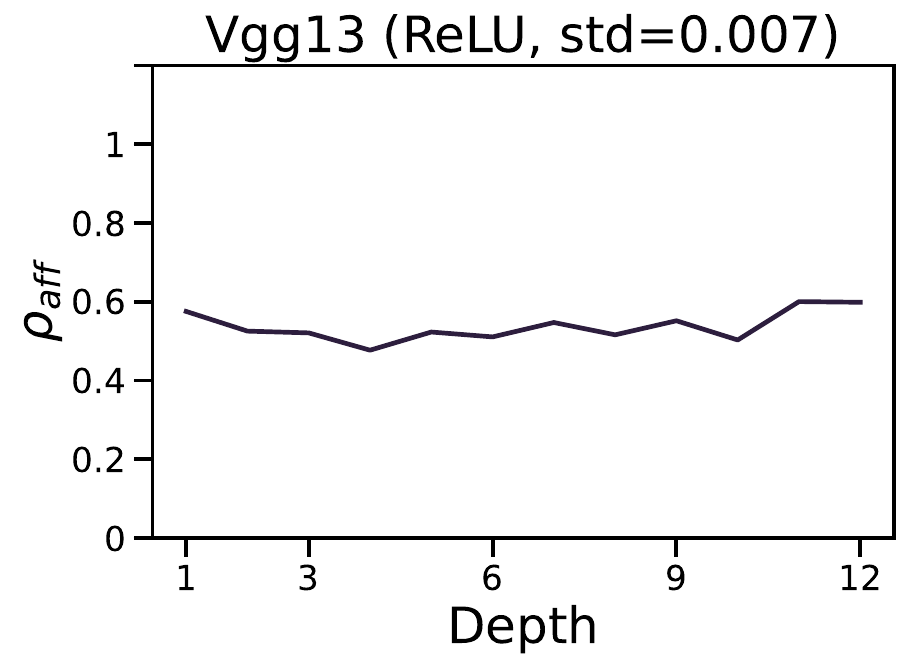}
    \includegraphics[width=.24\linewidth]{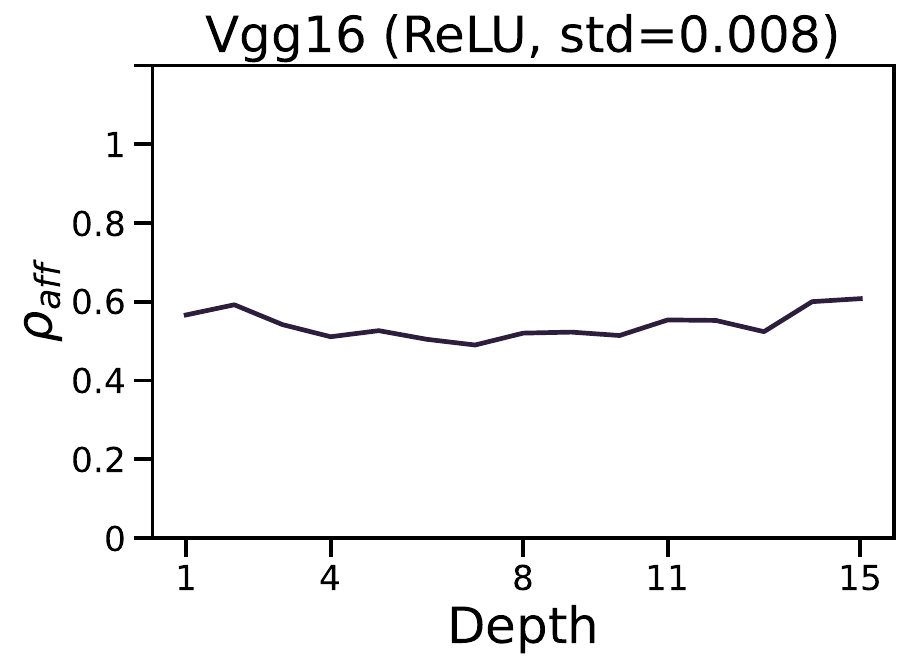}
    \includegraphics[width=.24\linewidth]{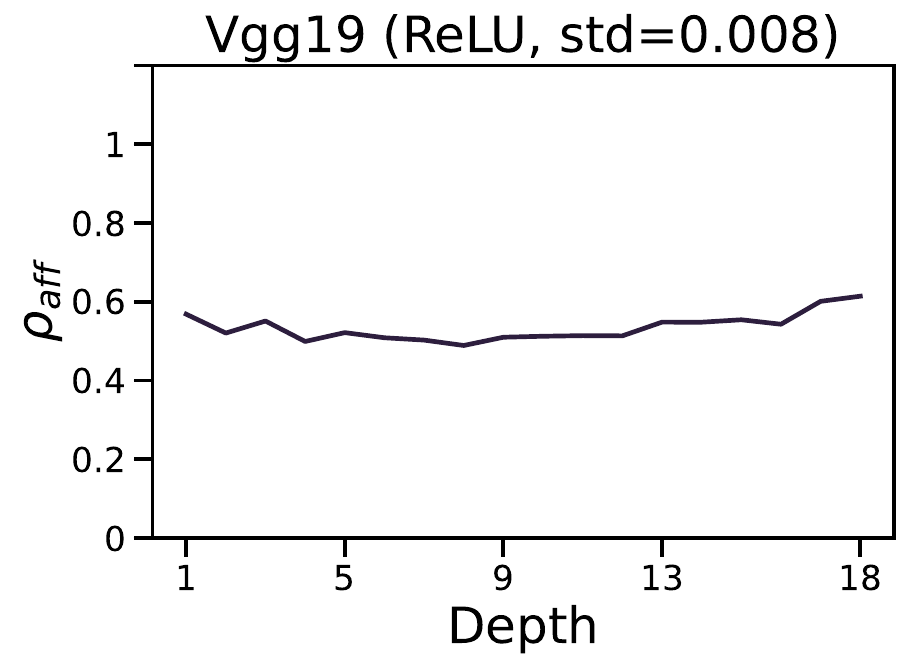}
    \caption{Impact of depth on the non-linearity signature of VGGs.}
    \label{fig:appendix_depth_vgg}
\end{figure*}
\begin{figure*}[t]
    \centering
    \includegraphics[width=.28\linewidth]{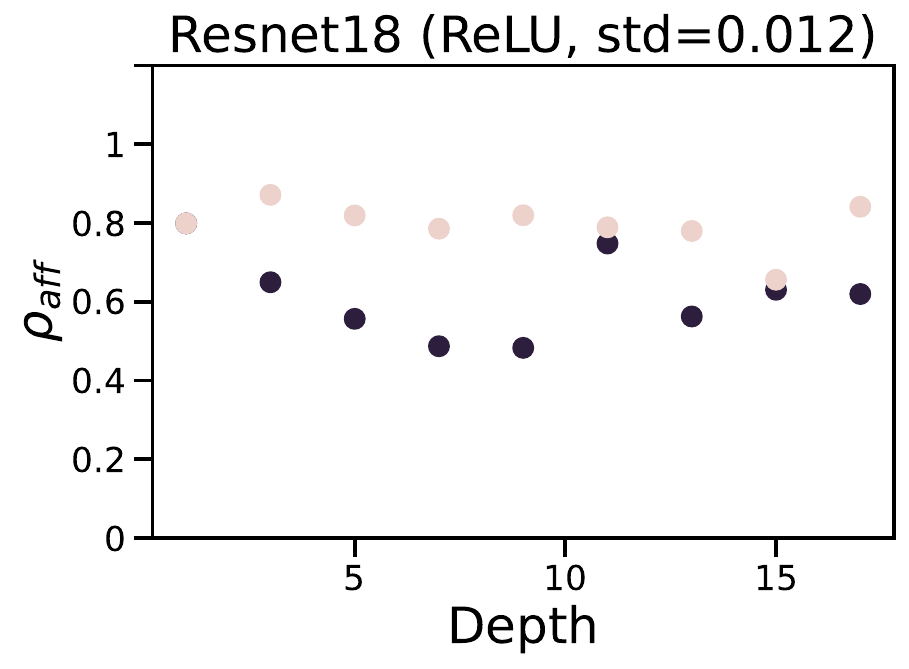}
    \includegraphics[width=.28\linewidth]{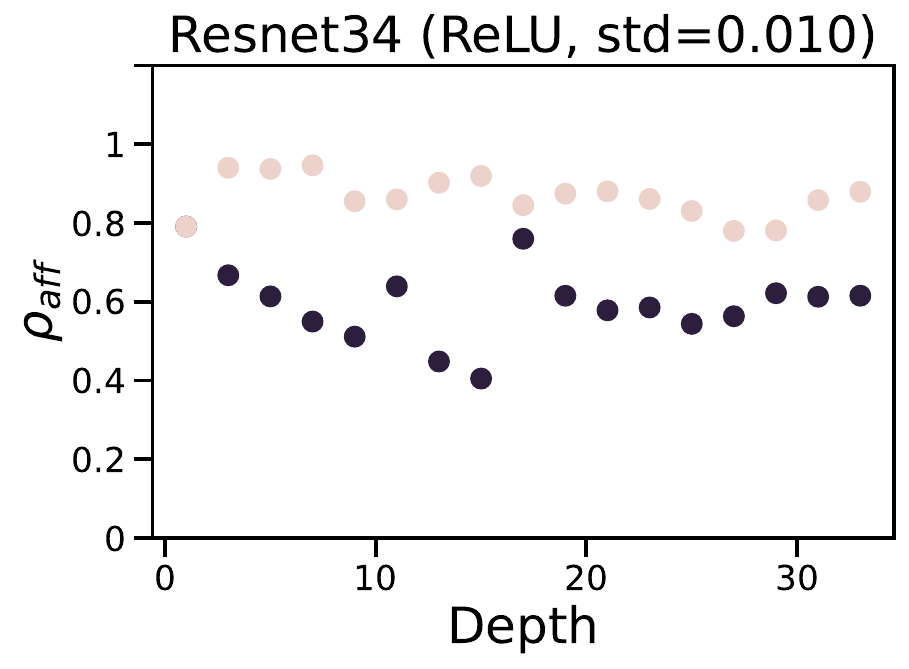}
    \includegraphics[width=.28\linewidth]{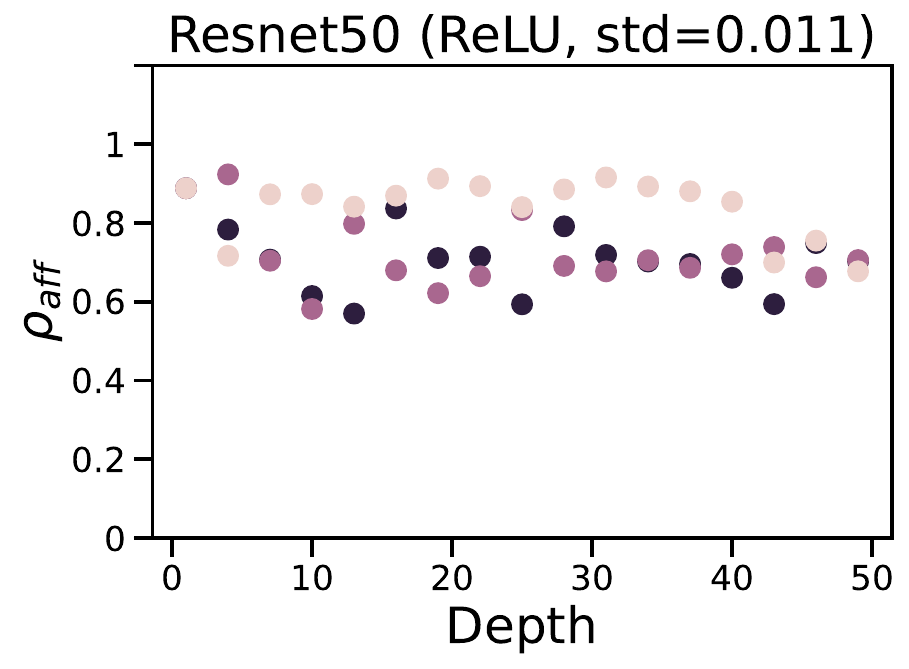}
    \includegraphics[width=.28\linewidth]{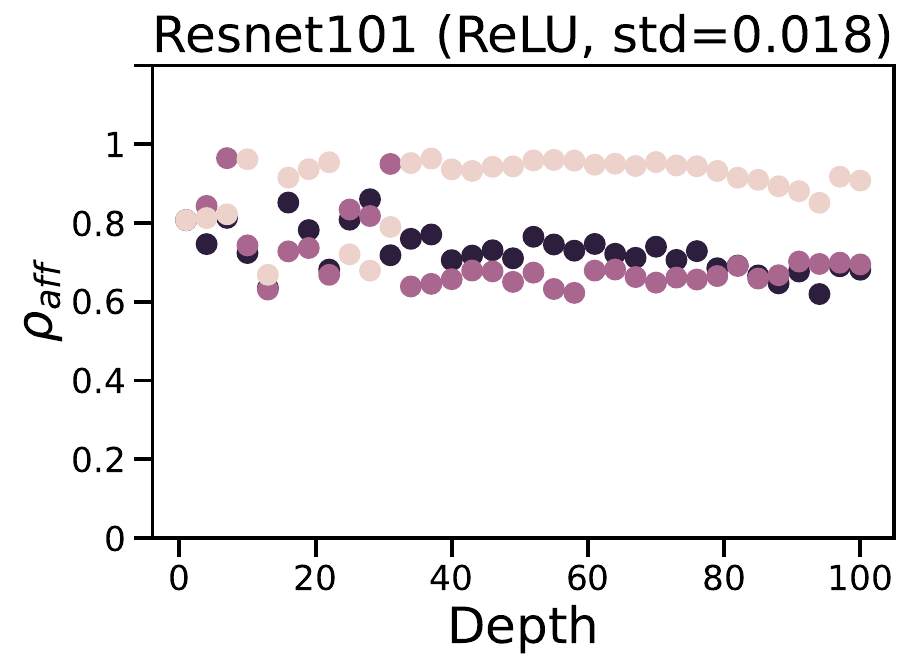}
    \includegraphics[width=.28\linewidth]{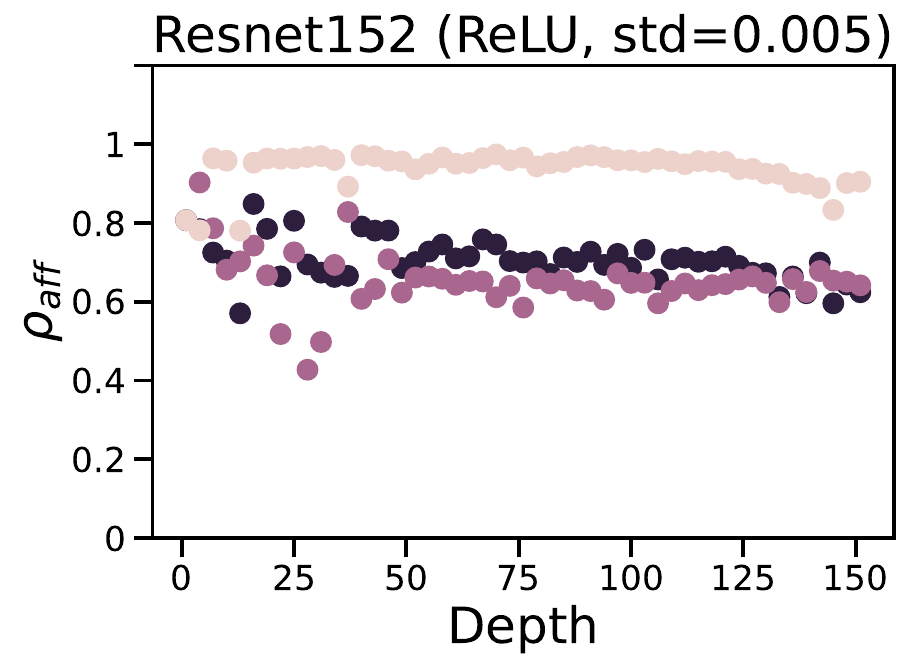}
    \caption{Impact of depth on the non-linearity signature of Resnets.}
    \label{fig:appendix_depth_resnet}
\end{figure*}

Finally, we explore the role of increasing depth for VGG and Resnet architectures. We consider VGG11, VGG13, VGG16 and VGG19 models in the first case, and Resnet18, Resnet34, Resnet50, Resnet101 and Resnet152. The results are presented in \cref{fig:appendix_depth_vgg} and \cref{fig:appendix_depth_resnet} for VGGs and Resnets, respectively.
Interestingly, VGGs do not change their non-linearity signature with increasing depth. In the case of Resnets, we can see that the separation between more linear post-residual activations becomes more distinct and approaches 1 for deeper networks.

\newpage
\section{Detailed comparisons between architectures}\label{ax:detailed_comp}
\begin{figure}[tb]
    \centering
    \includegraphics[width=\linewidth]{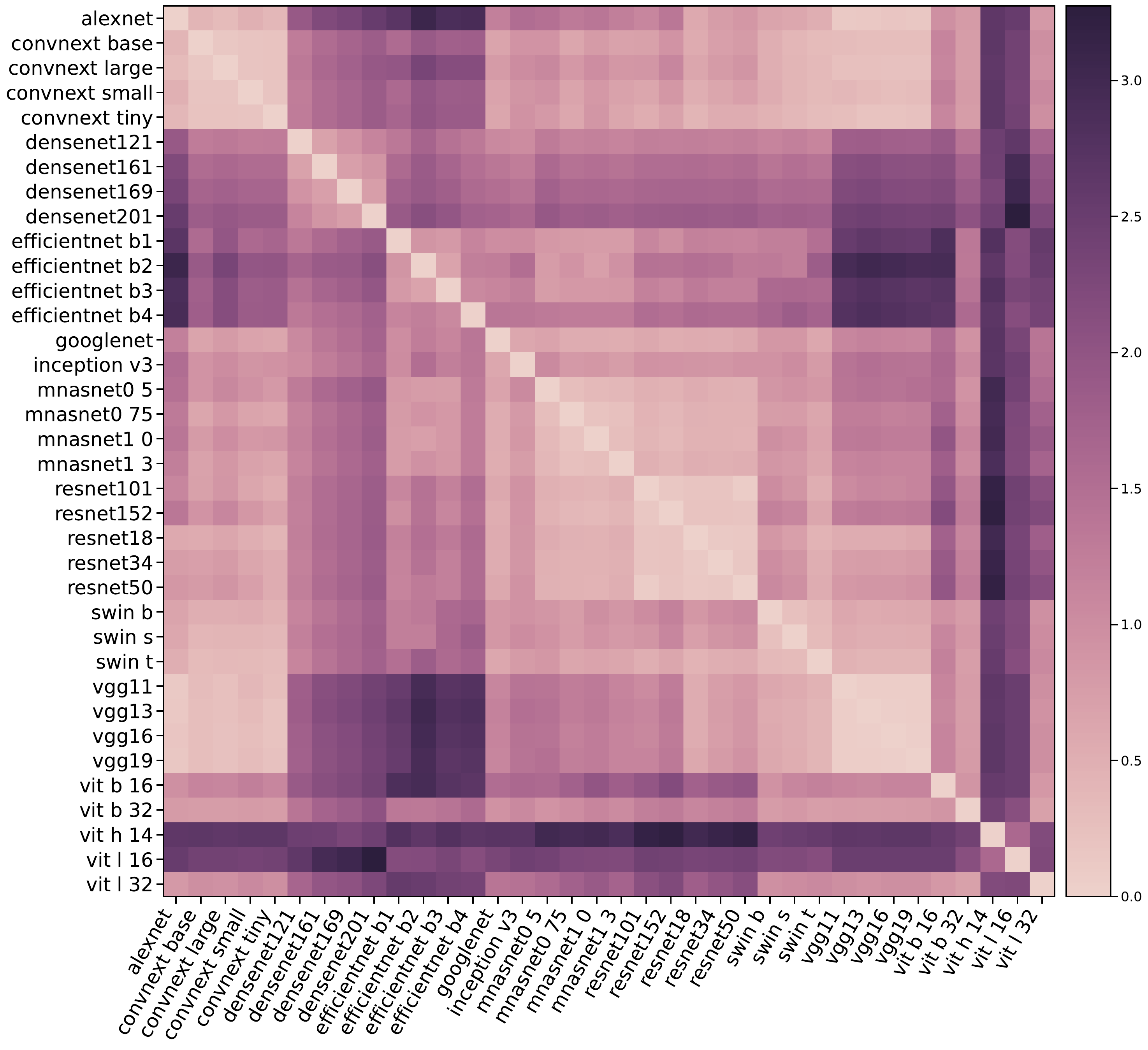}
    \caption{Full matrix of DTW distances between non-linearity signatures.}
    \label{fig:appendix_full_dtws}
\end{figure}

We consider the following metrics as 1) the linear \textsc{cka} \cite{kornblithCKA} commonly used to assess the similarity of neural representations, the average change in 2) \textsc{sparsity} and 3) \textsc{entropy} before and after the application of the activation function as well as the 4) Frobenius \textsc{norm} between the input and output of the activation functions, and the 5) $R^2$ score between the linear model fitted on the input and the output of the activation function. 
We present in \cref{tab:full_correlations}, the detailed values of Pearson correlations obtained for each architecture and all the metrics considered in this study. In \cref{fig:appendix_full_dtws}, we show the full matrix of pairwise DTW distances \cite{sakoe1978dynamic} obtained between architectures, then used to obtain the clustering presented alongside. For the latter, we applied multi-dimensional scaling algorithms to the linkage matrix of the 36 considered architectures.

\begin{table*}[tb]
    \centering
    \begin{tabular}{lccccc}
        \hline
        Model & CKA & Norm & Sparsity & Entropy & $R^2$ \\
        \hline
        alexnet & -0.75 & \textbf{-0.86} & 0.14 & -0.80 & -0.41 \\
        vgg11 & -0.07 & -0.76 & -0.15 & \textbf{-0.95} & -0.27 \\
        vgg13 & 0.08 & -0.66 & -0.23 & \textbf{-0.93} & -0.26 \\
        vgg16 & 0.01 & -0.63 & -0.19 & \textbf{-0.88} & -0.17 \\
        vgg19 & -0.01 & -0.62 & -0.15 & \textbf{-0.86} & -0.14 \\
        googlenet & 0.74 & -0.60 & \textbf{-0.83} & -0.49 & 0.73 \\
        inception v3 & 0.69 & -0.66 & \textbf{-0.75} & -0.45 & 0.35 \\
        resnet18 & 0.59 & -0.17 & \textbf{-0.67} & -0.30 & -0.44 \\
        resnet34 & 0.48 & -0.18 & \textbf{-0.65} & -0.19 & -0.08 \\
        resnet50 & 0.56 & -0.60 & -0.71 & -0.50 & \textbf{-0.78} \\
        resnet101 & 0.51 & -0.57 & \textbf{-0.70} & -0.51 & -0.64 \\
        resnet152 & 0.52 & -0.51 & \textbf{-0.68} & -0.42 & -0.48 \\
        densenet121 & 0.84 & -0.75 & \textbf{-0.87} & -0.62 & 0.82 \\
        densenet161 & \textbf{0.87} & -0.74 & \textbf{-0.87} & -0.67 & 0.81 \\
        densenet169 & \textbf{0.87} & -0.74 & \textbf{-0.87} & -0.67 & 0.81 \\
        densenet201 & 0.89 & -0.75 & \textbf{-0.91} & -0.67 & 0.90 \\
        efficientnet b1 & 0.35 & \textbf{-0.41} & -0.39 & 0.01 & 0.03 \\
        efficientnet b2 & \textbf{0.49} & -0.02 & -0.44 & -0.06 & 0.34 \\
        efficientnet b3 & \textbf{0.32} & -0.12 & -0.18 & -0.13 & 0.18 \\
        efficientnet b4 & 0.30 & \textbf{-0.51} & -0.29 & -0.44 & 0.11 \\
        vit b 32 & 0.47 & -0.31 & -0.29 & 0.39 & \textbf{0.51} \\
        vit l 32 & -0.14 & \textbf{-0.61} & -0.47 & -0.02 & -0.06 \\
        vit b 16 & -0.27 & \textbf{-0.71} & 0.04 & 0.39 & -0.22 \\
        vit l 16 & -0.39 & \textbf{-0.89} & -0.66 & -0.23 & -0.24 \\
        vit h 14 & -0.77 & -0.83 & \textbf{0.92} & 0.31 & -0.49 \\
        swin t & -0.12 & -0.39 & -0.02 & \textbf{-0.42} & -0.06 \\
        swin s & -0.003 & \textbf{-0.61} & -0.31 & 0.18 & -0.03 \\
        swin b & -0.32 & \textbf{-0.59} & -0.43 & 0.42 & -0.32 \\
        convnext tiny & 0.77 & -0.01 & -0.04 & 0.09 & \textbf{0.80} \\
        convnext small & 0.57 & 0.22 & 0.25 & 0.13 & \textbf{0.72} \\
        convnext base & 0.67 & 0.41 & 0.35 & -0.03 & \textbf{0.82} \\
        convnext large & 0.75 & 0.23 & 0.35 & -0.10 & \textbf{0.84} \\
        \hline
        Average & 0.31 $\pm$ 0.45 & \textbf{-0.44} $\pm$ \textbf{0.35} & -0.31 $\pm$ 0.43 & -0.29 $\pm$ 0.39 & 0.13 $\pm$ 0.50 \\
    \end{tabular}
    \caption{Pearson correlations between the affinity score and other metrics, for all the architectures evaluated in this study. We see that no other metric can reliably provide the same information as the proposed non-linearity signature across different neural architectures.}
    \label{tab:full_correlations}
\end{table*}

\begin{figure}[tb]
    \centering
    \includegraphics[width=\linewidth]{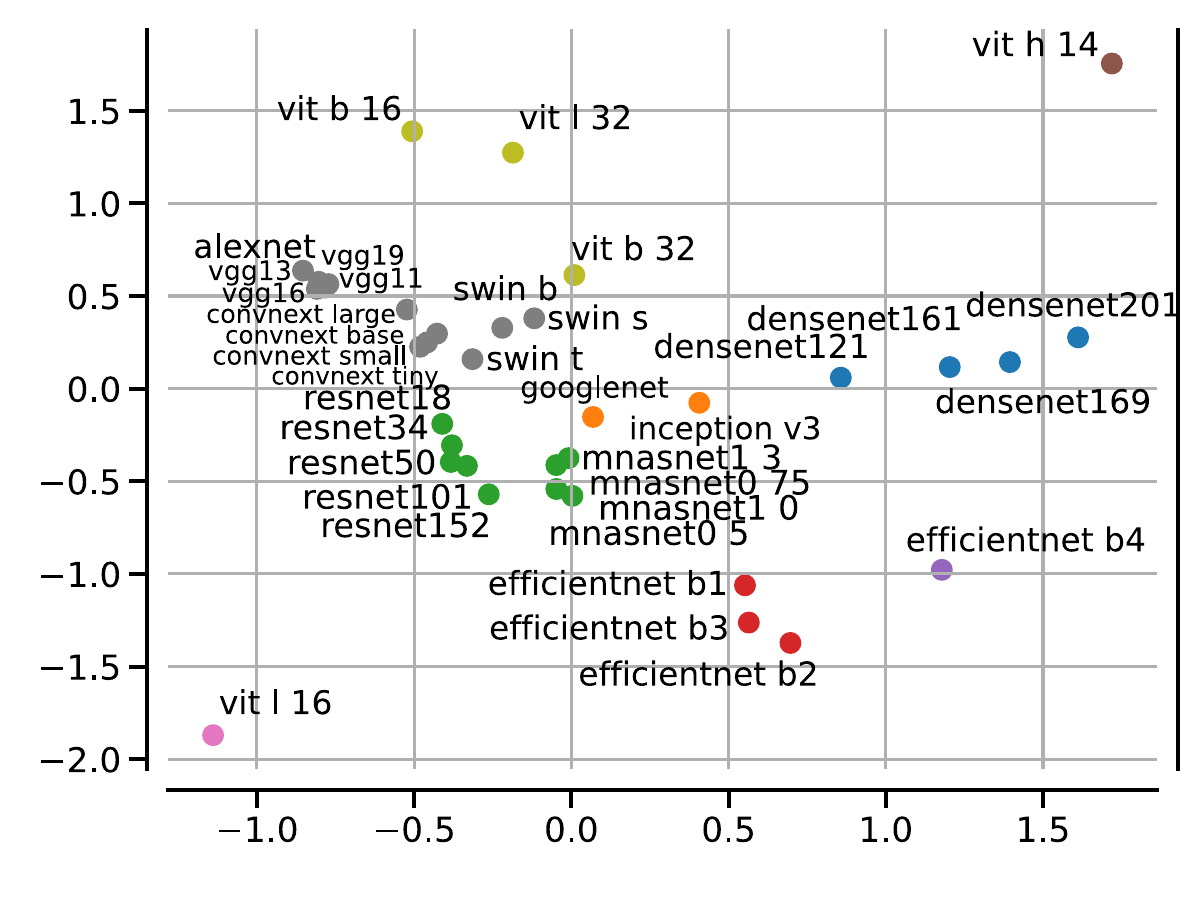}
    \caption{Multi-dimensional scaling of the linkage matrix obtained from the pairwise DTW distances between non-linearity signatures.}
    \label{fig:appendix_mds}
\end{figure}

\newpage
\section{Results on more datasets}
\label{sec:appendix_datasets}
\begin{figure*}
    \centering
    \includegraphics[width=.85\linewidth]{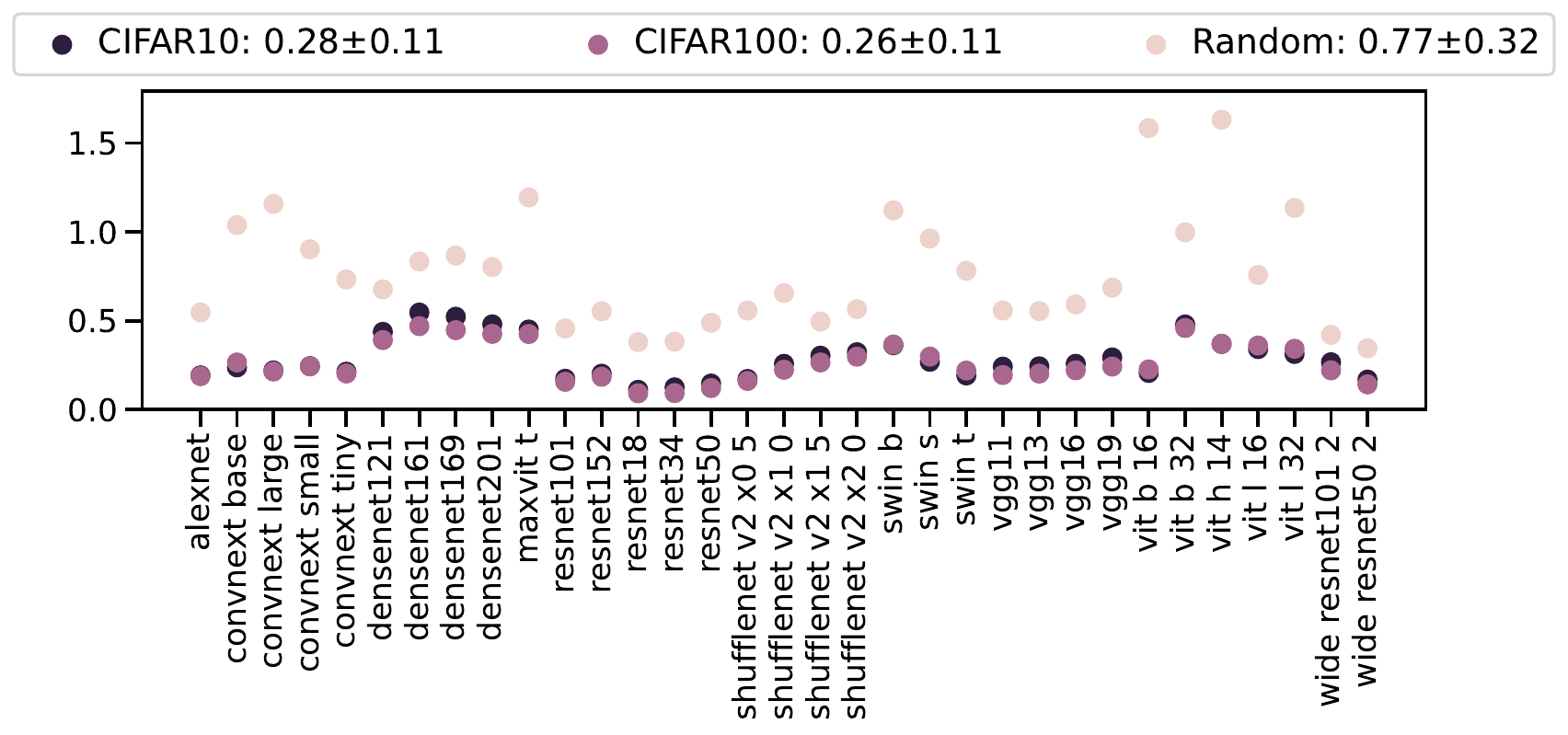}
    \caption{Deviation in terms of the Euclidean distance of the non-linearity signature obtained on CIFAR10, CIFAR100, and Random datasets from the non-linearity signature of the Imagenet dataset.}
    \label{fig:appendix_datasets_dtws}
\end{figure*}

Below, we compare the results obtained on CIFAR10, CIFAR100 datasets as well as when the random data tensors are passed through the network. As the number of plots for all chosen 33 models on these datasets will not allow for a meaningful visual analysis, we rather plot the differences  -- in terms of the DTW distance -- between the non-linearity signature of the model on Imagenet dataset with respect to three other datasets. We present the obtained results in Figure \ref{fig:appendix_datasets_dtws}.

We can see that the overall deviation for CIFAR10 and CIFAR100 remains lower than for Random dataset suggesting that these datasets are semantically closer to Imagenet.

\newpage
\section{Results for self-supervised methods}
\label{sec:appendix_ssl}
\begin{figure*}[tb]
    \centering
    \includegraphics[width=.9\linewidth]{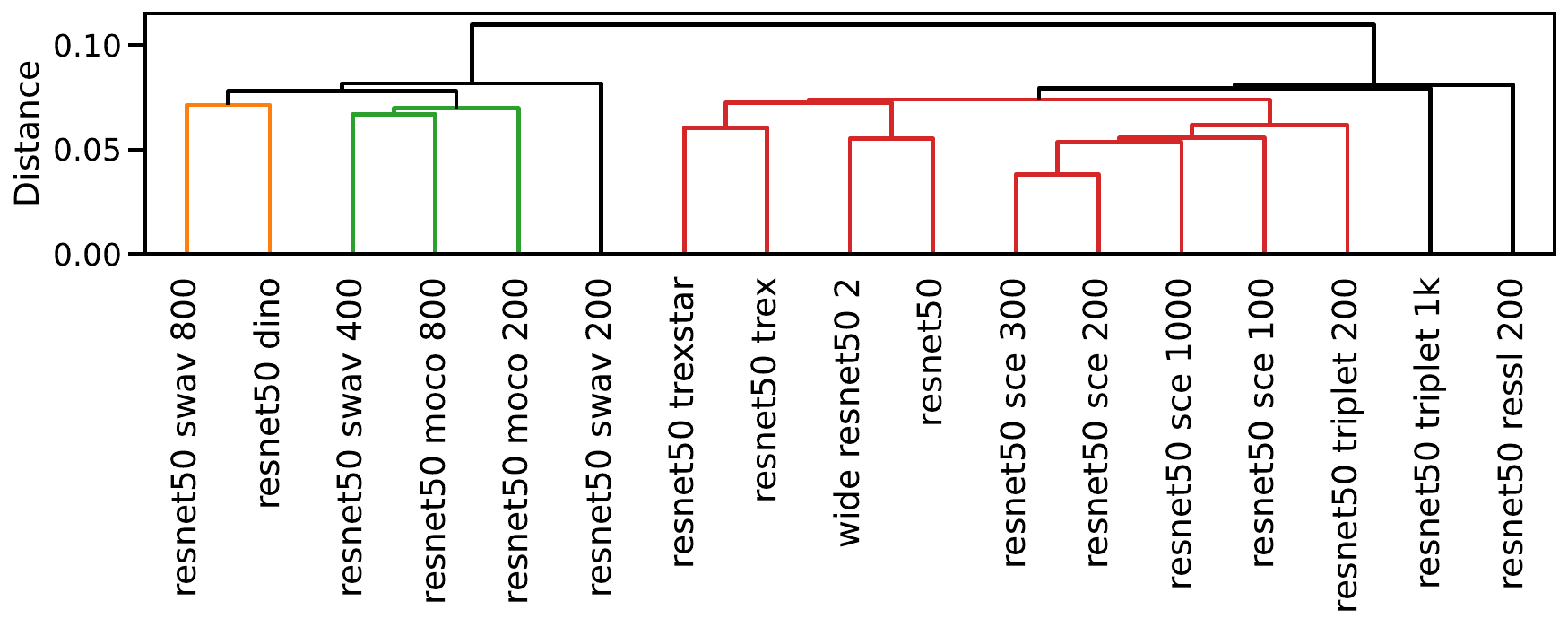}
    \caption{Hierarchical clustering of supervised and self-supervised pre-trained Resnet50 using the DTW distances between their non-linearity signatures.}
    \label{fig:clustering_ssl_sup}
\end{figure*}

In this section, we show that the non-linearity signature of a network remains almost unchanged when considering other pertaining methodologies such as for instance, self-supervised ones. To this end, we use 17 Resnet50 architecture pre-trained on Imagenet within the next 3 families of learning approaches:
\begin{enumerate}
    \item SwAV \citep{caron2020unsupervised}, DINO \citep{caron2021emerging}, and MoCo \citep{he2020momentum} that belong to the family of contrastive learning methods with prototypes;
    \item Resnet50 \citep{he2016deep}, Wide Resnet50 \citep{zagoruyko2016wide}, TRex, and TRex* \citep{sariyildiz2023no} that are supervised learning approaches;
    \item SCE \citep{denize2023similarity}, Truncated Triplet \citep{wang2021solving}, and ReSSL \citep{zheng2021ressl} that perform contrastive learning using relational information.
\end{enumerate}
From the dendrogram presented in Figure \ref{fig:clustering_ssl_sup}, we can observe that the DTW distances between the non-linearity signatures of all the learning methodologies described above allow us to correctly cluster them into meaningful groups. This is rather striking as the DTW distances between the different instances of the Resnet50 model are rather small in magnitude suggesting that the affinity scores still retain the fact that it is the same model being trained in many different ways. 

\begin{figure}[tb]
    \centering
    \includegraphics[width=\linewidth]{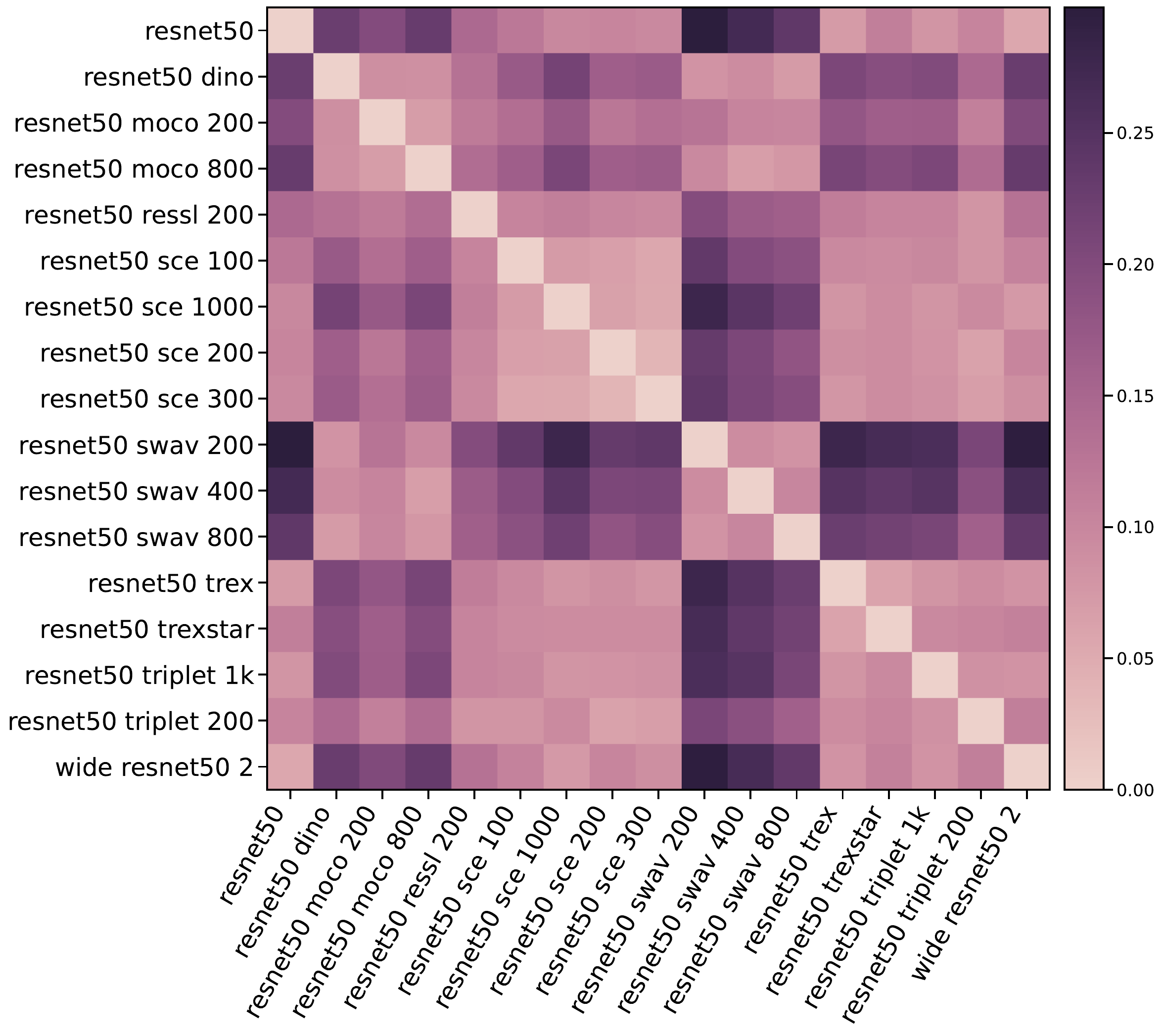}
    \caption{DTW distances associated with the clustering presented in \cref{fig:clustering_ssl_sup}. We can see distinct clusters as revealed by the dendrogram.}
    \label{fig:dtws_resnet50}
\end{figure}

\renewcommand{\arraystretch}{1.15}
\begin{table}[tb]
    \centering
    \begin{tabular}{l|c}
        Criterion & Mean $\pm$ std \\
        \hline
      $\rho_\text{aff}$   &  0.76$\pm$\textbf{0.04} \\
      Linear CKA   & 0.90$\pm$0.07\\
      Norm & 448.56$\pm$404.61\\
      Sparsity & 0.56$\pm$0.16\\
      Entropy  & 0.39$\pm$0.46
    \end{tabular}
    \caption{Robustness of the different criteria when considering the same architectures pre-trained for different tasks. Affinity score achieves the lowest standard deviation suggesting that it is capable of correctly identifying the architecture even when it was trained differently.}
    \label{tab:aff_score_resnet50}
\end{table} 

While providing a fine-grained clustering of different pre-trained models for a given fixed architecture, the average affinity scores over batches remain surprisingly concentrated as shown in \cref{tab:aff_score_resnet50}. This hints at the fact that the non-linearity signature is characteristic of architecture but can also be subtly multi-faceted when it comes to its different variations.